\pdfoutput=1
\documentclass[11pt]{article}
\usepackage[table]{xcolor}

\usepackage[final]{acl}
\usepackage{graphicx}
\usepackage{subcaption}
\usepackage{multirow}
\usepackage{times}
\usepackage{latexsym}
\usepackage{tabularx}
\usepackage{booktabs}
\usepackage[T1]{fontenc}
\usepackage{microtype}

\usepackage{inconsolata}

\usepackage{graphicx}

\usepackage{booktabs}

\usepackage{tikz}
\usepackage{enumitem}
\usepackage{array}
\usepackage{pifont}
\newcommand{\badge}[2]{%
\tikz[baseline]{\node[rounded corners=2pt, fill=#1!15, draw=#1!60, inner xsep=4pt, inner ysep=2pt]
{\small\textbf{#2}};}}
\definecolor{Set2BlueGreen}{HTML}{66C2A5}
\definecolor{Set2Orange}{HTML}{FC8D62}
\definecolor{Set2PurpleBlue}{HTML}{8DA0CB}
\newcommand{\best}{\cellcolor{blue!30}}
\newcommand{\secondbest}{\cellcolor{blue!10}}

\definecolor{blue1}{RGB}{234, 230, 255} 
\definecolor{blue2}{RGB}{194, 200, 255} 
\definecolor{blue3}{RGB}{154, 170, 255}

\definecolor{red1}{RGB}{255,245,238}
\definecolor{red2}{RGB}{255,228,225}
\definecolor{red3}{RGB}{255,188,185}

\definecolor{light-gray}{HTML}{E5E4E2}
\definecolor{light-cyan}{HTML}{E0FFFF}
\newcolumntype{R}{>{\raggedleft\arraybackslash}p{0.28cm}}
\newcolumntype{L}{>{\raggedleft\arraybackslash}p{0.23cm}}

\title{Annotating Dimensions of Social Perception in Text: \\A Sentence-Level Dataset of Warmth and Competence}

\author{Mutaz Ayesh$^1$, Saif M. Mohammad$^2$, Nedjma Ousidhoum$^1$\\
$^1$ Cardiff University, $^2$ National Research Council Canada \\
Correspondence: \texttt{OusidhoumN@cardiff.ac.uk}
}

\begin{document}
\maketitle
\begin{abstract}
\textit{Warmth} (W) (often further broken down into \textit{Trust} (T) and \textit{Sociability} (S)) and \textit{Competence} (C) are central dimensions along which people evaluate individuals and social groups \cite{fiske-endure-2018}. While these constructs are well established in social psychology, they are only starting to get attention in NLP research through word-level lexicons, which do not fully capture their contextual expression in larger text units and discourse. In this work, we introduce \textit{Warmth and Competence Sentences (W\&C-Sent)}, the first sentence-level dataset annotated for warmth and competence. The dataset includes over 1,600 English sentence--target pairs annotated along three dimensions: \textit{trust} and \textit{sociability} (components of \textit{warmth}), and \textit{competence}\footnote{Available at \url{https://github.com/nedjmaou/W_C_Sent}}. The sentences in W\&C-Sent are social media posts that express attitudes and opinions about specific individuals or social groups (the targets of our annotations). We describe the data collection, annotation, and quality-control procedures in detail, and evaluate a range of large language models (LLMs) on their ability to identify trust, sociability, and competence in text. W\&C-Sent provides a new resource for analyzing warmth and competence in language and supports future research at the intersection of NLP and computational social science.

%\medskip
\textcolor{red}{\noindent\textbf{Content warning:} This paper contains potentially offensive examples.}

\end{abstract}

\section{Introduction}

Language is a powerful medium through which people express their emotions and opinions about topics and individuals. In social psychology, interpersonal evaluation is largely structured around warmth and competence, the two primary dimensions along which people form impressions of, and make judgments about, individuals and social groups \cite{Fiske2002,fiske-endure-2018}. To explain how such evaluations manifest, the Stereotype Content Model (SCM), proposed by \citet{Fiske2002}, examines social perception, bias, and stereotyping. Central to the SCM is the idea that humans rapidly infer whether others are well-intentioned and socially oriented---even in the case of strangers---before assessing their ability to act on those intentions. Accordingly, \textbf{warmth (W)} reflects perceptions of \textbf{trust (T)} and \textbf{sociability (S)}, capturing whether an individual is seen as benevolent, cooperative, and inclined toward positive social interaction. In contrast, \textbf{competence (C)} reflects perceived capability to act on those intentions.
These dimensions are fundamental to the study of social interaction, emotional responses, stereotypes, and human behavior more broadly. As such, understanding them is crucial for building human-centered NLP and AI systems. 

However, while the study of warmth and competence is well established in psychology, they have only recently begun to receive attention in NLP. 
For example, \citet{wordsofwarmth} introduced a lexicon of over 42k words and multiword expressions associated with warmth, sociability, trust, and competence. 
Lexical approaches to language analysis are simple yet powerful, and are especially useful for identifying aggregate trends (e.g., across large collections of social media posts over time) \cite{teodorescu-mohammad-2023-evaluating}. Nevertheless, for applications that require instance-level understanding (e.g., determining the warmth communicated in a sentence), machine learning approaches trained on sentence-level annotated datasets are markedly more accurate.
One reason is that meaning does not correspond to the simple sum of individual words' senses \cite{Szabo2024Compositionality}. Instead, expressions of trust, sociability, and competence are often shaped by syntax, compositional semantics, and pragmatic cues, which cannot be fully captured at the word level. Consequently, word-based resources provide limited insight into how these social traits are conveyed in context. 

To address this gap, we build W\&C-Sent---a dataset designed to capture contextualized expressions of trust, sociability, and competence toward specific targets at the sentence level.
We collect instances from the SemEval-2016 stance dataset \cite{StanceSemEval2016}---chosen for its focus on opinions about specific targets---and augment them with instances from the Affect, Body, Cognition, Demographic, and Emotion (ABCDE) dataset \cite{wahle-etal-2026-abcde}. Each instance is then independently annotated by 4 to 7 fluent English speakers for each dimension (T, S, and C) with respect to seven distinct targets: individual politicians (Hillary Clinton, Barack Obama, and Donald Trump) and social groups (women, religious people, atheists, and climate change activists). Annotations are recorded on a 7-point scale ranging from -3 (very low) to +3 (very high), with 0 representing neutrality (e.g., see \autoref{fig:badge-examples}). 

We provide a detailed description of our data collection, annotation, and quality-control procedures and make the dataset publicly available. Then, we evaluate a range of LLMs on their ability to detect trust, sociability, and competence in text. We find that LLMs struggle to reliably assess these social dimensions, leading to suboptimal performance in downstream applications and their deployment in real-world systems, such as chatbots, content moderation, and machine translation. W\&C-Sent supports a broad range of applications in NLP and social science including analytics, annotation, discourse analysis, and bias studies. It also serves as a benchmark for evaluating whether models accurately capture trust, sociability, and competence in language.
 
\begin{figure}[t]
\centering
\small
\begin{minipage}[t]{0.41\textwidth}
\centering
\fbox{
\begin{minipage}{\linewidth}
\textbf{\texttt{Target:}} \texttt{Hillary Clinton}\\
\textbf{\texttt{Text:}} \texttt{Would you wanna be in a long term relationship with some bitch that hides her emails, \& lies to your face? Then \#Dontvote} \\[4pt]
\badge{red}{\texttt{High Distrust -3}}\quad
\badge{Set2Orange}{\texttt{Moderate Unsociability -2}}\quad
\badge{Set2BlueGreen}{\texttt{Moderate Competence +2}}
\end{minipage}}
\end{minipage}
\hfill
\begin{minipage}[t]{0.41\textwidth}
\centering
\fbox{
\begin{minipage}{\linewidth}
\textbf{\texttt{Target:}} \texttt{Religious people}\\
\textbf{\texttt{Text:}} \texttt{Could all those who believe in a god please leave. The meeting will now continue for the grown ups only.} \\[4pt]
\badge{yellow}{\texttt{Neutral trust 0}}\quad
\badge{red}{\texttt{High Unsociability -3}}\quad
\badge{red}{\texttt{High Incompetence -3}}
\end{minipage}}
\end{minipage}

\caption{Examples of sentence--target pairs with different annotations across trust, sociability, and competence.}
\label{fig:badge-examples}
\vspace*{-3mm}
\end{figure}

\section{Related work}

Prior work in social perception has established warmth and competence as core dimensions underlying social judgments. The Stereotype Content Model \citep{Fiske2002} showed that different combinations of warmth and competence elicit distinct emotional responses (e.g., pity or envy), explaining why groups are associated with varied affective reactions rather than uniform prejudice \citep{fiske-endure-2018}. Subsequent work by \citet{agency-communion} challenged a unidimensional view of warmth and proposed a multifaceted conceptualization of warmth (communion), distinguishing between morality-related traits (e.g., honesty, fairness) and sociability-related traits (e.g., friendliness, empathy). They further demonstrated that morality plays a stronger role in group judgments. This distinction was further supported by \citet{koch-morality}, who showed that these facets cluster differently and predict several behavioral outcomes, motivating more fine-grained modeling of warmth and competence.
This framework is particularly helpful for assessing bias and stereotypes in NLP \citep{fraser-etal-2024-stereotype}, given the growing interest in bias and stereotype evaluation in NLP models \citep{kiritchenko-mohammad-2018-examining, blodgett-etal-2020-language, stereoset, ousidhoum-etal-2021-probing}, and in LLMs in particular \citep{cheng-etal-2023-marked, siddique-etal-2024-better, plaza-del-arco-etal-2024-angry, plaza-del-arco-etal-2024-divine}.

Current computational approaches to warmth and competence are largely lexicon-based. The Valence--Arousal--Dominance (VAD) lexicon \citep{vad-2018, nrcvadlexiconv2} provides large-scale word-level scores for valence---previously argued to be the largest component of warmth \citep{Cuddy2007TheBM}---and dominance (also referred to as competence), offering broad coverage but treating warmth as a single dimension. Other work has developed small dictionaries specifically for warmth and competence text analysis \citep{nicolas-dictionary}. \citet{wordsofwarmth} introduced the first large-scale lexicon, Words of Warmth, providing reliable human ratings of word--warmth, word--sociability, word--trust, and word--competence associations for over 42k words and multiword expressions.\footnote{\url{https://saifmohammad.com/WebPages/warmth.html}}
\citet{wordsofwarmth} used the lexicon to study: (1)\ the rate at which children acquire warmth-, ~sociability-, and trust-related words in their vocabulary with age; and (2)\ the average degree of warmth and competence of words that co-occur with mentions of social groups (such as ``immigrant'', ``doctor'', ``girl'', etc.) in social media.
The Words of Warmth lexicon provides a foundation for more nuanced modeling of social traits in language and motivates extending such analyses beyond the word level, which we address in this paper.

\section{Dataset}

\begin{figure*}[!ht]
\centering
\small
\fbox{
\begin{minipage}{0.68\textwidth}

\textbf{\texttt{Instance from the SemEval-2016 Stance Dataset:} \newline}\\ 
\colorbox{lightgray}{
\begin{minipage}{0.95\textwidth}
\texttt{We need Obama out and @realDonaldTrump in the White House ASAP.}
\end{minipage}
}\\[8pt]
\begin{tabular}{p{0.47\textwidth} | p{0.47\textwidth}}
\textbf{\texttt{SemEval-2016 Labels}} & \textbf{ $\rightarrow$ \texttt{W\&C-Sent Labels}} \\[2pt]
\midrule
\textbf{\texttt{Original Target:}} \newline \badge{Set2PurpleBlue} {\texttt{Donald Trump}} & 
\textbf{$\rightarrow$ \texttt{Extracted Target:}} \newline \badge{blue} {\texttt{Barack Obama}} \\[2pt]
\textbf{\texttt{Original Label:}} \newline \badge{Set2PurpleBlue}{\texttt{Stance: In favor}} &
\textbf{$\rightarrow$ \texttt{Social Perception Labels:}} \newline
\badge{Set2Orange}{\texttt{C: Moderate Incompetence}}\quad \newline
\badge{Set2Orange}{\texttt{S: Slight Unsociability}}\quad \newline
\badge{Set2Orange}{\texttt{T: Moderate Distrust}} \\\end{tabular}

\end{minipage}
}
\caption{An example illustrating how the SemEval-2016 Stance dataset was used to extract sentence--target pairs for W\&C-Sent. Specifically, a social media post is initially labeled as expressing a stance in favor of Donald Trump; however, because it also mentions Barack Obama, we additionally extract Barack Obama as a target and annotate it for competence, sociability, and trust (C, S, and T).}
\label{fig:repurpose-stance-to-wc}
\vspace*{-0.3mm}
\end{figure*}

We construct a dataset of 1,633 sentences annotated for warmth (W)---broken down into trust (T) and sociability (S)---and competence (C). We describe the data sources, annotation procedures, and quality-control processes in the following.

\subsection{Data Sources}

\paragraph{Stance Data}
We use the SemEval-2016 Stance dataset \citep{StanceSemEval2016} as the primary source of textual instances (1,490 sentences; 90.8\%). It consists of social media posts manually annotated for stance toward predefined targets. Specifically, each post is labeled according to whether the expressed stance toward a given target is \textit{in favor of}, \textit{against}, or \textit{neutral} (more precisely, neither favor nor against stance can be inferred).
 The text may express stance overtly or implicitly. 
 The dataset also includes labels for the overall sentiment of the post independent of stance. 
 The six predefined targets in the dataset are \textit{Hillary Clinton}, \textit{Donald Trump}, \textit{Atheism}, \textit{Feminism}, \textit{Climate Change}, and \textit{Abortion}.

\paragraph{The ABCDE Dataset}
We additionally use the ABCDE dataset \citep{wahle-etal-2026-abcde}, a large-scale corpus of over 400 million text utterances from diverse sources, including social media, blogs, and books. The dataset contains self-reported features that capture affective, cognitive, linguistic, bodily, and demographic information.
We selected 151 sentences from the ABCDE dataset, constituting 9.2\% of the total sentences in \textit{W\&C-Sent}, using a simple keyword-based search targeting the seven predefined targets. The search keywords can be found in Appendix \ref{appendix:abcde-selection}.

\paragraph{Data Selection}
\autoref{tab:summary-stat} presents the final distribution of the dataset.
Human targets (e.g., \textit{Hillary Clinton}, \textit{Donald Trump}, and \textit{women}) were prioritized, as the analysis of warmth and competence requires targets to be human individuals or social groups. Non-human or abstract targets---such as \textit{Atheism}, \textit{Feminism}, \textit{Climate Change}, and \textit{Abortion}---were included only when sentences explicitly referred to human subjects or collectives. Specifically, we retained posts discussing these topics and manually reassigned the targets to relevant groups of people rather than the topics themselves (e.g., \textit{atheism} posts referring to religious or non-religious people; \textit{abortion}-related posts referring to women; and \textit{climate change} posts targeting environmentalists or climate change activists) (see \autoref{fig:repurpose-stance-to-wc}).
Moreover, several instances in the SemEval stance dataset with \textit{Hillary Clinton} or \textit{Donald Trump} as targets contained references to \textit{Barack Obama}. In such cases, we created additional instances by pairing the same sentence with \textit{Barack Obama} as the target, resulting in a final set of seven targets.

\begin{table}[t]
\centering
\small
\begin{tabular}{lccc}\hline
\textbf{Target} & \textbf{SemEval} & \textbf{ABCDE} & \textbf{Total (\%)}  \\
\hline
Hillary Clinton         & 610               & 7              & 617 (37.7)\\

Donald Trump            & 301               & 108             & 409 (25.0)\\

Women                   & 292               & 21               & 313 (19.0)\\

Barack Obama            & 109               & 4                 & 113 (7.0)\\

Religious P.        & 98               &  7               & 105 (6.5)\\

Environ.       & 37             & 3                    & 40 (2.5)\\

Nonreligious P.      & 35            & 1                    & 36 (2.3)\\
\hline
\textbf{Total} & \textbf{1,482} & \textbf{151}  & \textbf{1,633 (100)}\\
\hline

\end{tabular}
\caption{\textbf{Distribution of targets in W\&C-Sent,} showing frequency counts by source dataset and overall percentages for the seven targets (Hillary Clinton, Donald Trump, women, Barack Obama, religious people (Religious P.), climate change activists (Environ.), and atheists (Non-religious P.).}
\label{tab:summary-stat}
\vspace*{-3mm}
\end{table}

\subsection{Annotation Process}

We preprocessed the data instances by anonymizing social media posts, replacing most \texttt{@mentions} with \texttt{@user}, except for mentions of public figures (e.g., \textit{Hillary Clinton}). The instances were then labeled using the annotation platform Prolific.

\paragraph{Annotator Selection Criteria}
Annotators were required to be fluent English speakers based in English-speaking countries and to have an approval rate above 99\% on Prolific. They were compensated at a rate of US \$16--22 per hour.

To reduce the likelihood of bot participation or inattentive responses, annotators were required to complete multiple attention checks throughout the study in which they were instructed to assign a specific, predetermined label (e.g., \textit{``Please assign a +3 score''}). Selecting any other label was treated as a failure to follow the instructions.

In addition, a set of potentially unambiguous sentences ($n=229$), manually pre-annotated by the authors was used in quality assessment. Annotators were presented with a random subset of these sentences, and if they failed to correctly annotate more than $20\%$ of them, their work was subject to full review and typically discarded.

\paragraph{Instructions}
To reduce annotator cognitive load, we annotate trust, sociability, and competence independently. We adapted the word-level guidelines proposed by \citet{wordsofwarmth} for sentence--target pair annotations.
The full annotation guidelines are provided in Appendix~\ref{appendix:english-guidelines}.
Annotators are shown a post together with a target and are asked to rate each of the three dimensions on a 7-point ordinal scale. Scores range from $-3$---very low trust, very low sociability, or very low competence---to $+3$---very high trust, very high sociability, or very high competence---depending on the dimension being annotated. A score of $0$ indicates neutrality, non-applicability, or a lack of (expressed) information.

\paragraph{Agreement and Reliability}
We compute average split-half reliability (SHR), a commonly used measure for assessing the reliability of ordinal-scale annotations \cite{Kuder1937TheTO,Cronbach1951CoefficientAA,Weir2005QuantifyingTR}. All annotations are randomly split into two halves, and separate aggregate scores are computed for each half. The similarity between the two sets of scores is then measured using a correlation metric. This procedure is repeated 1,000 times, and the resulting correlations are averaged \cite{wordsofwarmth}. The resulting SHR scores are $0.76$ for trust, $0.68$ for sociability, and $0.56$ for competence, indicating relatively high annotation reliability.
For inter-annotator agreement (IAA), we compute Krippendorff's $\alpha$, and obtain $0.60$ for trust, $0.50$ for sociability, and $0.30$ for competence, which is expected for subjective and fine-grained classification tasks. By contrast, when computing the overall average pairwise agreement (APA) for coarsened labels---mapping 
$[-3,-0.5]$ to \textit{low}, $[-0.5, 0.5]$ to \textit{neutral}, and $[0.5,3]$ to \textit{high}---the APA increases substantially, reaching $62.8\%$ for both trust and sociability, and $52.2\%$ for competence.

\subsection{Final Dataset}
\paragraph{Aggregated Scores and Statistics}

\begin{figure}
\centering
\includegraphics[width=0.9\linewidth]{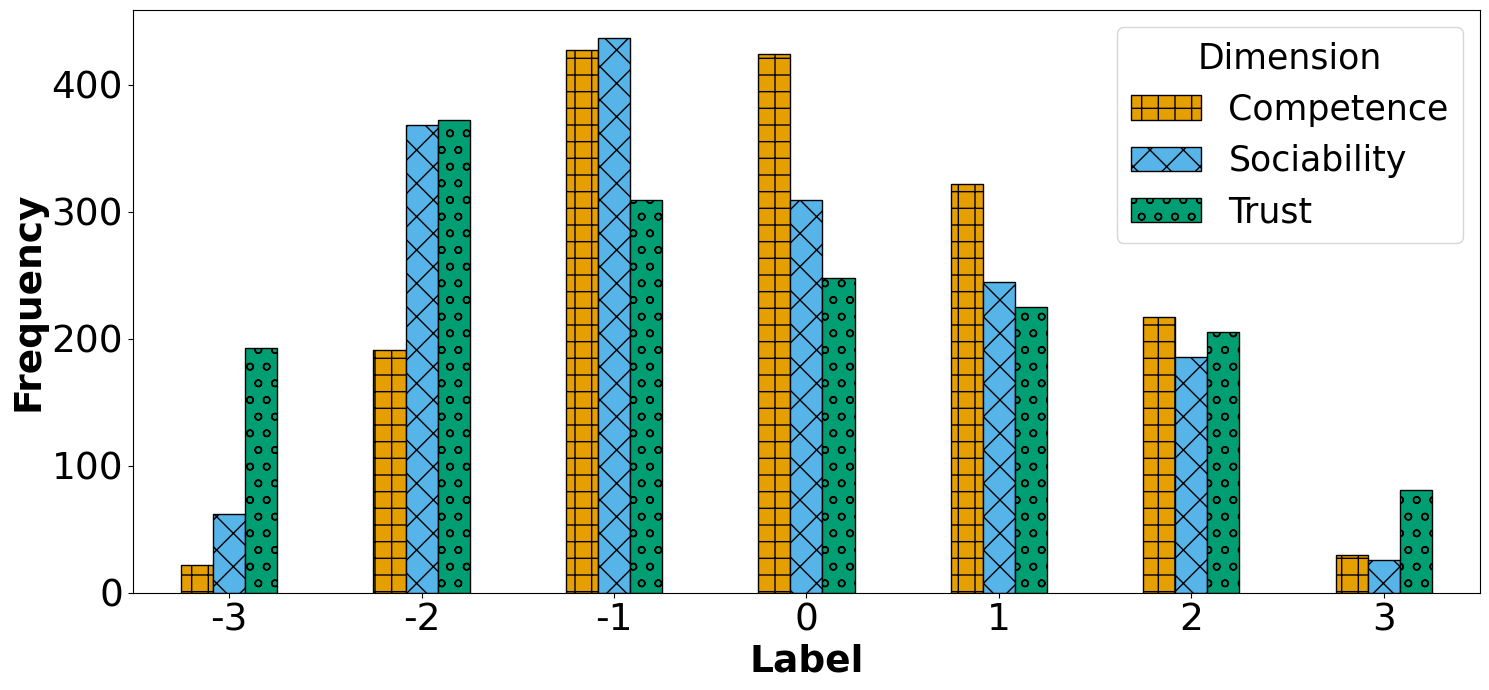}
\caption{Distribution of mean scores assigned to sentence--target pairs (1,633 pairs per dimension) in the dataset. The x-axis shows fine-grained labels, and the y-axis shows their frequency.}
\label{fig:mean-agg}
\vspace*{-3mm}
\end{figure}

To obtain a single label per sentence–target pair, we aggregate annotator scores using the mean, following \citet{wordsofwarmth}. \autoref{fig:mean-agg} shows the distribution of mean labels across the three dimensions, while \autoref{tab:coarse-labels} reports the distribution of these scores after coarsening them into three categories: Low (negative), Neutral, and High (positive). Further details are provided in Appendix \ref{appendix:coarse-procedure}.

The neutral label is the most frequent overall, particularly for competence. However, the dataset also contains a substantial proportion of non-neutral instances, with negative scores more prevalent than positive ones. This likely reflects the nature of the source data, as the original SemEval instances were drawn from discussions of controversial topics in the USA, such as elections and reproductive rights \cite{StanceSemEval2016}.

Across dimensions, sociability shows a concentration at -1 and -2 (low to moderate unsociability), while trust exhibits relatively high frequencies at -2 and especially -3 (moderate to high distrust). In contrast, positive labels (+1 to +3) are less frequent overall, with +3 being the rarest category. For competence and sociability, extreme values (+3 and -3) are similarly uncommon, suggesting that annotators tend to avoid assigning the most extreme ratings.

%\paragraph{Distribution of Coarse Labels}\label{subsubsec:coarsening}

\section{Experiments}

\subsection{Setup}

We split our dataset into 60\%/20\%/20\% for train, development, and test sets, and use our test set to evaluate several automatic baselines on the task of predicting continuous and ordinal trust/sociability/competence (T/S/C) scores (-3 to 3) for a given sentence--target pair:
\begin{enumerate}[label=--, noitemsep, nolistsep, leftmargin=*]
\item a majority-class (\textit{``dummy''}) baseline classifier,
\item a logistic regression (LR) classifier,
\item a fine-tuned BERTweet model \citep{bertweet}, and
\item zero-shot (ZS) and few-shot (FS) experiments with a range of open- and closed-source LLMs.
\end{enumerate}

\subsubsection{Baseline Models}

To predict continuous trust, sociability, and competence scores---each defined on a scale from -3 to +3---we train majority-vote (``dummy'') and logistic regression (LR) models using TF--IDF embeddings. Specifically, we train one model per dimension, resulting in six classifiers in total (i.e., three dummy and three LR models) that serve as baselines.

\subsubsection{Fine-tuned BERT}
We fine-tune three BERTweet-based classifiers on our training set to predict continuous trust, sociability, and competence scores, each defined on a scale from -3 to +3.

Additionally, we evaluate our baselines in a coarse-grained classification setting, where models predict whether the perceived trust, sociability, and competence of text--target pairs are \textit{``low''}, \textit{``neutral''}, or \textit{``high''}. As in the regression experiments, BERTweet models are fine-tuned separately for each dimension.
\begin{table}[!t]
\centering
\small
\resizebox{\columnwidth}{!}{%
\begin{tabular}{lrrr}\toprule
\textbf{Coarse label} & Low (\%)            & Neutral (\%)       & High (\%) \\ \midrule
\textbf{Trust}        & 945   (57.9)   & 105 (6.4)      & 583 (35.7) \\ 
\textbf{Sociability}  & 1,012  (62.0)   & 85  (5.2)      & 536 (32.8) \\
\textbf{Competence}   & 773  (47.3)    & 162  (9.9)     & 698 (42.7) \\ \bottomrule
\end{tabular}}
\caption{Distribution of W\&C-Sent's labels after coarsening. The table shows the strong prevalence of negative (low) labels compared to positive and neutral ones.}
\label{tab:coarse-labels}
\vspace*{-3mm}
\end{table}

\subsubsection{LLM Prompting}
\paragraph{Models}
We prompt six different LLMs: three open-weight and instruction-tuned (Gemma-3-4b \citep{gemma_2025}, Qwen-2.5-7b \citep{qwen2.5}, and Qwen3-4B-Instruct-2507 \cite{qwen3technicalreport}), and three closed-source models (Open AI's GPT-4o, GPT-4o-mini \citep{gpt4}, and GPT-5.2 Chat \cite{singh2025openai}). We design three separate prompts---one per dimension---in accordance with our annotation framework. That is, similar to the annotation process, in which each social dimension is labeled independently to reduce annotator's cognitive load, we evaluate each LLM on one dimension at a time. This design choice limits the amount of information included in each prompt and helps mitigate potential under-performance due to prompt overload \citep{lost-in-the-middle}. In addition, we prompted the models to assess sentence--target pairs using both fine-grained labels and coarse-grained labels, and in zero-shot (ZS) and few-shot (FS) settings.

\begin{table}[!ht]
\centering
\small
\resizebox{\columnwidth}{!}{%
\begin{tabular}{llccc}
\toprule
\textbf{Dimension} & \textbf{Model} & \textbf{Accuracy} & \textbf{F1} & \textbf{$\pm1$ Acc.} \\
\midrule
\multirow{15}{*}{\textbf{\texttt{Trust}}}
 & \texttt{Dummy}                    & 0.22 & 0.08 & 0.58 \\
 & \texttt{LR}                       & 0.30 & 0.30 & 0.65 \\
 & \texttt{Fine-tuned BERTweet}          & 0.35 & 0.31 & 0.83 \\
 & \texttt{Gemma3 ZS}                & 0.36 & 0.34  & 0.78  \\
 & \texttt{Gemma3 FS}                & 0.38 & 0.33 & 0.79 	   \\
 & \texttt{Qwen2.5 ZS}               & 0.27  & 0.25  & 0.72  \\
 & \texttt{Qwen2.5 FS}               & 0.35 	 & 	 0.35 	 & 	 0.76 	   \\
 & \texttt{Qwen3 ZS}                  & 0.32 & 0.30 & 0.78  \\
 & \texttt{Qwen3 FS}             & 0.30 & 0.26 & 0.78  \\ 
 & \texttt{GPT-4o ZS}                & \best\textbf{0.43}  & \best\textbf{0.42}  & \best\textbf{0.91}  \\
 & \texttt{GPT-4o FS}                & 0.39 	 & 	\secondbest0.40 	 & 	 0.84 	   \\
 & \texttt{GPT-4o-mini ZS}           & 0.27  & 0.26  & 0.60  \\
 & \texttt{GPT-4o-mini FS}           & 0.20 	 & 	 0.17 	 & 	 0.54  \\
& \texttt{GPT-5.2 ZS}                & \secondbest0.41 & 0.38 & 0.87  \\
& \texttt{GPT-5.2 FS}            & 0.40 & 0.39 & \secondbest0.89  \\
\midrule
\multirow{15}{*}{\textbf{\texttt{Sociability}}}
 & \texttt{Dummy}                    & 0.26 & 0.11 & 0.67 \\
 & \texttt{LR}                       & 0.31 & 0.26 & 0.70 \\
 & \texttt{Fine-tuned BERTweet}          & \best\textbf{0.46} & 0.34 & \secondbest0.88 \\
 & \texttt{Gemma3 ZS}                & 0.34  & 0.27  & 0.82  \\
 & \texttt{Gemma3 FS}                & 0.31 	 & 	 0.23 	 & 	 0.76 	   \\
 & \texttt{Qwen2.5 ZS}               & 0.20  & 0.20 & 0.69  \\
 & \texttt{Qwen2.5 FS}               & 0.25 	 & 	 0.25 	 & 	 0.67 	 \\
 & \texttt{Qwen3 ZS}                  & 0.35 & 0.30 & 0.81 \\
 & \texttt{Qwen3 FS}              & 0.31 & 0.24 & 0.82  \\
 & \texttt{GPT-4o ZS}                & \secondbest0.44 & \best\textbf{0.40}& \best\textbf{0.92}  \\
 & \texttt{GPT-4o FS}                & 0.38 	 & 	 0.35 	 & 	 0.87 	   \\
 & \texttt{GPT-4o-mini ZS}           & 0.13  & 0.14 & 0.52  \\
 & \texttt{GPT-4o-mini FS}           & 0.24 	 & 	 0.20 	 & 	 0.59  \\
& \texttt{GPT-5.2 ZS}               & 0.31 & 0.30 & 0.83  \\
& \texttt{GPT-5.2 FS}           & 0.40 & \secondbest0.37 & \secondbest0.88  \\
\midrule
\multirow{15}{*}{\textbf{\texttt{Competence}}}
 & \texttt{Dummy}                    & 0.24 & 0.09 & 0.60 \\
 & \texttt{LR}                       & 0.19 & 0.16 & 0.62 \\
 & \texttt{Fine-tuned BERTweet}          & \best\textbf{0.36} & 0.24 & \best\textbf{0.86}\\
 & \texttt{Gemma3 ZS}                & 0.22  & 0.19 & 0.58  \\
 & \texttt{Gemma3 FS}                & \secondbest0.35 	 & 	 0.27 	 & 	 0.70  \\
 & \texttt{Qwen2.5 ZS}               & 0.22  & 0.19  & 0.59  \\
 & \texttt{Qwen2.5 FS}               & 0.30 	 & 	 \secondbest0.28 	 & 	 0.65 	   \\
 & \texttt{Qwen3 ZS}                  & 0.22 & 0.18 & 0.60\\
 & \texttt{Qwen3 FS}             & 0.25 & 0.21 & 0.67  \\
 & \texttt{GPT-4o ZS}                & 0.34 & \best\textbf{0.31} &\secondbest0.80\\
 & \texttt{GPT-4o FS}                & 0.24 	 & 	 0.24 	 & 	 0.64 	   \\
 & \texttt{GPT-4o-mini ZS}           & 0.09  & 0.11  & 0.37  \\
 & \texttt{GPT-4o-mini FS}           & 0.22 	 & 	 0.20 	 & 	 0.60 	  \\
& \texttt{GPT-5.2 ZS}                & 0.28 & 0.25 & 0.73 \\
& \texttt{GPT-5.2 FS}            & 0.27 & 0.24 & 0.71 \\
\bottomrule
\end{tabular}}
\caption{Fine-grained classification performance of different baselines---majority-class (\textit{dummy}), logistic regression (\textit{LR}), and fine-tuned BERTweet---along with LLM models in zero-shot (ZS) and few-shot (FS) settings for \textbf{trust}, \textbf{sociability}, and \textbf{competence}. Scores are reported on the test set. The \textbf{best score} for each metric and dimension is shown in \textbf{bold} and highlighted in blue, and the second-best is highlighted in a lighter shade of blue.}
\label{tab:fine-grained-results}
\end{table}

\paragraph{Prompts}

Each prompt starts with a definition of the targeted social dimension (i.e., sociability, trust, or competence). It then establishes analytical constraints by instructing the model to interpret the text snippet as an average person would and to evaluate the social dimension at the sentence level. The prompts also include definitions of pragmatic phenomena common in social media---such as sarcasm, hyperbole, and irony---as disclaimers to mitigate errors observed in pilot experiments, in which LLMs interpreted sarcastic or ironic sentences literally.
The target entity is explicitly specified, and the model is instructed to evaluate each sentence with respect to that entity and its definitions, while ignoring references to other individuals or social groups, if present.

For each dimension, the instructions closely follow the annotation guidelines provided to human annotators. This section specifies the label set that the model is expected to use, ensuring that categorical labels are returned in both fine-grained settings (e.g., ``high distrust'', ``moderate sociability'', ``slight incompetence'') and coarse-grained ones (``low'', ``neutral'', or ``high''). In zero-shot prompts, examples are omitted, whereas few-shot prompts include examples covering all labels (8–9 examples for fine-grained prompts and 5 for coarse-grained prompts). To activate the models' reasoning capabilities, the prompts instruct the LLMs to reason first and then return the label. All prompts are provided in Appendix~\ref{appendix:llm-prompts}.

\subsection{Experimental Results}

Tables \ref{tab:fine-grained-results} and \ref{tab:coarse-grained-results} present the performance of all fine-grained and coarse-grained classification models, respectively. We report \textbf{accuracy} and the \textbf{weighted F1 score}, as well as the \textbf{within-1 ($\pm1$) accuracy} for the fine-grained classifiers. Given the \textbf{ordinal nature} of the task, $\pm1$ accuracy indicates how often model predictions fall within one rating level of the true labels, thereby capturing near-miss performance. Precision and recall are reported in the Appendix (Tables \ref{tab:fine-prec-recall} and \ref{tab:coarse-prec-recall}).

\paragraph{Fine-grained Classification Results}

\autoref{tab:fine-grained-results} presents the performance of the different baselines on the fine-grained classification task. As expected, the majority-class (dummy) classifier performs the worst, followed by logistic regression (LR). However, LLMs do not perform substantially better in either zero-shot or few-shot settings, with no model exceeding 50\% accuracy. This is particularly noticeable for the competence dimension, which appears to be the most challenging---even for human annotators.

In terms of accuracy, fine-tuned BERTweet performs best on sociability and competence, followed by GPT-4o (ZS) on sociability and Gemma3 (FS) on competence. GPT-4o (ZS) achieves the highest accuracy on trust, followed by GPT-5.2. In terms of weighted F1, GPT-4o (ZS) outperforms all other models, with its few-shot counterpart close behind, highlighting the difficulty of the task and the test set for LLMs. Gemma3 outperforms the GPT and Qwen models in the few-shot setting on competence, achieving an accuracy of 0.35. In contrast, Qwen2.5 and GPT-4-mini sometimes fail to outperform the dummy classifier or the LR model, indicating their limitations for this task.

The $\pm1$ accuracy scores suggest that GPT-4o (ZS) tends to make more subtle errors than other models, achieving 0.91 and 0.92 on trust and sociability, respectively. It is outperformed only by the fine-tuned BERTweet model on competence (0.86), where it remains a close second (0.80). Overall, larger and more recent closed models (e.g., GPT-5.2) do not necessarily outperform older ones.
\begin{table}[!ht]
\centering
\small
\resizebox{\columnwidth}{!}{%
\begin{tabular}{clcc}
\toprule
\textbf{Dimension} & \textbf{Model} & \textbf{Accuracy} & \textbf{F1} \\
\midrule
\multirow{5}{*}{\textbf{\texttt{Trust}}}
 & \texttt{Fine-tuned BERTweet} & 0.79 & 0.59  \\
 & \texttt{Gemma3 ZS}         & 0.74  & 0.60  \\
 & \texttt{Gemma3 FS}         & 0.72  & 0.60  \\
 & \texttt{Qwen2.5 ZS}          & 0.62  & 0.53  \\
 & \texttt{Qwen2.5 FS}          & 0.61  & 0.51  \\
& \texttt{Qwen3 ZS} & 0.79 	 & 	 0.66 \\
 & \texttt{Qwen3 FS} & 0.72 	 & 	 0.62 \\
 & \texttt{GPT-4o ZS}       & \best\textbf{0.86} & \best\textbf{0.78} \\
 & \texttt{GPT-4o FS}       & 0.77 & 0.71 \\
 & \texttt{GPT-4o-mini ZS}  & 0.78 & 0.65 \\
 & \texttt{GPT-4o-mini FS} & 0.70 	& 0.58 \\
 & \texttt{GPT-5.2 ZS} & \secondbest0.81 	 & 	 \secondbest0.72 \\
 & \texttt{GPT-5.2 FS} & 0.78 	 & 	 0.71 \\
\midrule
\multirow{5}{*}{\textbf{\texttt{Sociability}}}
 & \texttt{Fine-tuned BERTweet} & \secondbest0.81 & 0.54 \\
 & \texttt{Gemma3 ZS}     & 0.69  & 0.56  \\
 & \texttt{Gemma3 FS}     & 0.71  & 0.54  \\
 & \texttt{Qwen2.5 ZS}      & 0.47  & 0.42  \\
 & \texttt{Qwen2.5 FS}      & 0.34  & 0.34  \\
 & \texttt{Qwen3 ZS} & 0.74 	 & 	 0.59 	\\
 & \texttt{Qwen3 FS} & 0.67 	 & 	 0.58 	\\
 & \texttt{GPT-4o ZS}    & 0.80 & 0.71 \\
 & \texttt{GPT-4o FS}    & 0.72 & 0.66 	\\
 & \texttt{GPT-4o-mini ZS} & 0.74 & 0.54 \\
 & \texttt{GPT-4o-mini FS} & 0.62 & 0.47 \\
 & \texttt{GPT-5.2 ZS} & \best \textbf{0.83} 	 & 	 \best \textbf{0.74} 	 \\
 & \texttt{GPT-5.2 FS} & 0.79 	 & 	 \secondbest0.72 	 \\
\midrule
\multirow{5}{*}{\textbf{\texttt{Competence}}}
 & \texttt{Fine-tuned BERTweet} & \best \textbf{0.74} & 0.50 \\
 & \texttt{Gemma3 ZS}     & 0.52  & 0.46  \\
 & \texttt{Gemma3 FS}     & 0.52  & 0.43  \\
 & \texttt{Qwen2.5 ZS}      & 0.47  & 0.47  \\
 & \texttt{Qwen2.5 FS}      & 0.56  & 0.53 \\
  & \texttt{Qwen3 ZS} & 0.64 	 & 	 0.56 	\\
 & \texttt{Qwen3 FS} & \secondbest 0.67 	 & 	 \secondbest0.58 	 \\
 & \texttt{GPT-4o ZS}    & 0.63 & \best\textbf{0.59} \\
 & \texttt{GPT-4o FS}    & 0.60 & 0.57 \\
 & \texttt{GPT-4o-mini ZS} & 0.53 & 0.40 \\
 & \texttt{GPT-4o-mini FS} & 0.47 & 0.37 \\
 & \texttt{GPT-5.2 ZS} &  0.49 	 & 	 0.46 \\
 & \texttt{GPT-5.2 FS} & 0.46 	 & 	 0.43 \\
\bottomrule
\end{tabular}
}
\caption{Coarse-grained classification performance of fine-tuned BERT and LLM models in zero-shot (ZS) and few-shot (FS) settings for \textbf{trust}, \textbf{sociability}, and \textbf{competence}. Scores are reported on the test set. The \textbf{best score} for each metric and dimension is in \textbf{bold} and highlighted in blue, and the second-best is highlighted in a lighter shade of blue.}
\label{tab:coarse-grained-results}
\vspace*{-3mm}
\end{table}

\paragraph{Coarse-grained Classification Results}
The results for predicting coarse-grained labels, shown in Table \ref{tab:coarse-grained-results}, are higher than those for the fine-grained setting, with competence remaining the most challenging dimension. 

For the trust dimension, GPT-4o (ZS) achieves the highest accuracy and F1 scores, followed by GPT-5.2. The remaining models---except for Gemma3---perform similarly, with accuracy scores in the 70–79\% range. 
For sociability, GPT-5.2 (ZS) outperforms all other models on both metrics, followed by the fine-tuned BERTweet model in accuracy and GPT-5.2 (FS) in F1. Finally, for competence, the fine-tuned BERTweet model achieves the highest accuracy, while GPT-4o (ZS) attains the highest F1 score; Qwen3 is a close second on both metrics.

We further observe that, although performance improves considerably when labels are coarsened---similar to $\pm1$ accuracy in the fine-grained setting (see \autoref{tab:fine-grained-results})---the fine-tuned BERTweet model continues to perform consistently, whereas models such as Gemma3 and Qwen2.5 may still perform poorly. Notably, GPT-5.2 performs comparatively poorly in both zero-shot and few-shot settings on the coarsened labels, particularly for the competence dimension, despite its recency and scale. This finding suggests that fine-tuning a pretrained model on high-quality data can sometimes be more effective than relying on an LLM for challenging and subjective tasks.

\paragraph{Few-shot vs. Zero-shot Performance}
We observe that few-shot (FS) settings consistently underperform zero-shot (ZS) settings in this task. This pattern appears in Table \ref{tab:fine-grained-results} across several ZS–FS pairs for trust (GPT-4o, GPT-4o-mini, GPT-5.2, and Qwen3), sociability (Gemma3, GPT-4o, Qwen3), and competence (GPT-4o, GPT-5.2).
The coarse-grained results in Table \ref{tab:coarse-grained-results} show even more noticeable declines, particularly for trust in the Qwen3, GPT-4o, and GPT-4o-mini pairs, and for sociability in the GPT-4o, GPT-4o-mini, Qwen2.5, and Qwen3 pairs.

We analyze 102 instances in which FS predictions from both GPT-4o and Qwen2.5 substantially deviated from both gold labels and ZS outputs. Our findings suggest that models prompted in few-shot settings may amplify surface-level cues---such as all-caps text and exclamation marks. This effect is particularly evident in the frequent default to neutral predictions (32 of 39 cases for GPT-4o and 63 cases for Qwen2.5).
In addition, FS settings often misinterpret sarcasm or figurative language as literal meaning, by assigning neutral labels to ironic posts or over-penalizing sarcastic expressions (e.g., references to \textit{Hillary Clinton's ``sniper fire''}).
Overall, these results suggest that performance degradation in few-shot settings may be due to overly conservative calibration boundaries (i.e., calibration bias), therefore reducing sensitivity to pragmatic nuances.

\section{Analysis}
\begin{table}[!t]
\centering
\resizebox{\columnwidth}{!}{%
\begin{tabular}{lcccc}
\toprule
\textbf{Stance} & \textbf{Count} & \textbf{Trust}
 & \textbf{Sociability} & \textbf{Competence} \\ 
\midrule
\hfil Against & 561 &  -2 (-1.4 $\pm$ 1.4) &  -1 (-1.2 $\pm$ 1.1) &  -1 (-0.5 $\pm$ 1.2) \\ %\hline
\hfil Favor   & 262 &  2 (1.7 $\pm$ 0.9) &  1 (1.2 $\pm$ 0.9)  &  2 (1.4 $\pm$ 0.8) \\ %\hline
\hfil Neutral &  31 &  0 (-0.4 $\pm$ 0.7) & 0 (-0.2 $\pm$ 0.9)  & 0 (-0.3 $\pm$ 0.7) \\ 
\bottomrule
\end{tabular}}
\caption{Aggregated median values (mean $\pm$ standard deviation) for each stance subset from the SemEval-2016 stance dataset across the three dimensions. These statistics are computed over the 854 text--target pairs shared between W\&C-Sent and the SemEval-2016 Stance dataset.}
\label{tab:stance-dimensions}
\vspace*{-3mm}
\end{table}

\paragraph{To what extent do stance and sentiment correlate with perceptions of warmth and competence?}

We aim to understand how trust, sociability, and competence relate to stance and sentiment. That is, how people perceive a target individual or social group when they are in favor of or against a topic, and when they express positive or negative sentiment in language. As W\&C-Sent was primarily constructed using a stance dataset, 854 sentences (52.3\%) share targets with the SemEval-2016 Stance dataset. Investigating the relationship between each dimension and stance allows us to determine which traits most strongly influence stance, the role each dimension plays, and how stance is reflected through warmth (i.e., trust and sociability) and competence in language.  

\autoref{tab:stance-dimensions} shows how stance relates to assessments of trust, sociability, and competence. Sentences labeled as \textit{against} consistently receive negative median scores across all dimensions, with particularly low trust (-2). By contrast, sentences labeled as \textit{in favor} show positive median scores across all dimensions, including strong trust and competence scores (2), while \textit{neutral} sentences remain near zero with comparatively low variance. This pattern indicates that stance strongly conditions how targets are perceived, both in the direction and intensity of judgment.

\autoref{tab:spearman-correlations} provides further evidence, as the correlation coefficients indicate strong positive monotonic associations between stance and each dimension. The strongest correlation is observed for trust ($\rho = 0.71$), followed by sociability ($\rho = 0.69$) and competence ($\rho = 0.61$). The SemEval-2016 stance dataset was also annotated for sentiment, which follows a similar trend, with slightly higher correlation coefficients for trust and sociability.

\begin{table}[!t]
\centering
\small
\begin{tabular}{lccc}\toprule
                    & \textbf{Trust} & \textbf{Sociability} & \textbf{Competence} \\ \midrule
\textbf{Stance}    & 0.71           & 0.69                & 0.61 \\ 
\textbf{Sentiment} & 0.72           & 0.73                & 0.61 \\ \bottomrule
\end{tabular}
\caption{Spearman correlation coefficients between stance or sentiment and the median scores of trust, sociability, and competence. The results show strong positive monotonic associations, meaning that as stance or sentiment becomes more positive, the corresponding dimension scores also tend to increase.}
\vspace*{-3mm}
\label{tab:spearman-correlations}
\end{table}

\paragraph{When do annotators show high agreement or marked disagreement?} 

Annotators assigned the same fine-grained label in only 3\% of the dataset instances ($n=143$), whereas agreement on the same polarity (low, neutral, or high) occurred in approximately 26.9\% of cases ($n=1,316$ instances), with the majority reflecting negative judgments (see examples in \autoref{tab:unanimity-examples-fine}). Interestingly, we notice that agreement is higher for assessments of trust toward individual targets (e.g., Donald Trump and Barack Obama), while it is higher for assessments of sociability toward social groups (e.g., women and religious people). This might be linked to the nature of each dimension, as trust tends to relate to the personal and moral aspects of a given target, whereas sociability reflects its relational and social characteristics.

At the same time, we observe meaningful disagreement across dimensions on 159 sentence--target pairs whose aggregated scores show different signs across dimensions. \autoref{tab:div-comp} highlights a subset of these cases, in which negative trust and sociability scores strongly contrast with positive competence scores. Such instances demonstrate notable adherence to the annotation guidelines: even when a target is described in harsh or derogatory terms, annotators consistently consider the ability, skills, or power that the author seems to attribute to the target.
\begin{table}[!t]
\centering
% \small
\resizebox{\columnwidth}{!}{%
\begin{tabular}{p{1.5cm}p{5cm}cc}
\toprule
\textbf{Target} & \hfil \textbf{Text} & \textbf{Dim.} & \textbf{Label} \\
\midrule
Women & $@$readyforHRC $@$HillaryClinton \#HillaryClinton, the US presidency is a testament to the success of \#women their role in the world & C & +3 \\\hline

Barack Obama & I hate Hillary Clinton and Obama, please go die together thrown into the ocean get ripped into prices by sharks. & S & -3 \\ \hline

Donald Trump & ,$@$realDonaldTrump trumps presidential dreams r about as real as KimJonguns unicorns. & C & -3 \\ \bottomrule

\end{tabular}}
\caption{Examples of W\&C-Sent instances with full agreement on fine-grained labels across different dimensions (Dim.).}
\vspace*{-3mm}
\label{tab:unanimity-examples-fine}
\end{table}

\paragraph{To what extent do trust, sociability, and competence correlate with each other?}

It is worth noting that largely distinct sets of annotators participated in labeling each dimension. Therefore, when calculating correlations between the aggregated scores, the resulting values reflect relationships between the dimensions at a broad, population-level perceptual scale, rather than consistency in the judgments of individual annotators. In other words, these correlations capture general patterns in how people perceive trust, sociability, and competence, rather than the personal biases or preferences of the annotators.
%level, 

\autoref{tab:spearman-correlations-pairs} presents the correlations between each pair of dimensions using Spearman's $\rho$. The coefficients show moderately to strongly positive correlations among all three dimensions. Trust correlates most strongly with sociability ($\rho = 0.79$), which is expected, as both trust and sociability are commonly considered sub-dimensions of warmth. Trust correlates somewhat less strongly with competence ($\rho = 0.67$), while sociability and competence exhibit a similar level of correlation ($\rho = 0.68$). Overall, these values indicate that the three dimensions (T, S, and C) are interrelated in human perception. For instance, a person perceived as trustworthy is often also viewed as sociable to some degree, and individuals perceived as incompetent are frequently also seen as untrustworthy, and vice versa.
\begin{table}[!t]
\centering
\small
\resizebox{\columnwidth}{!}{%
\begin{tabular}{m{2cm}p{3cm}ccc}
\toprule
\textbf{Target} & \hfil \textbf{Sentence} & \textbf{T} & \textbf{S} & \textbf{C} \\
% \hline
% Donald Trump & $@$user $@$user The only reason Trump wants war is to distract the American people from the crimes he and his lapdogs have and are still committing . & -3 & -3 & 1 \\
\midrule
Hillary Clinton & .$@$HillaryClinton blames her lack of trust among the populace on $@$GOP, forgetting that she's a lying, conniving, murderous, cheat. & -3 & -3 & +1 \\ \hline
% 3 & Donald Trump & $@$user $@$user $@$realDonaldTrump Where do you get the information suggesting tRump has given away one third of his fortune since taking office . From what I can see he has used the office of President to enrich himself and his cronies . & -3 & -2 & 2 \\
Donald Trump & If there was such a thing as the ``American'' Hitler. Donald Trump would probably be that guy. \#nationalism & -3 & -3 & +2 \\
\bottomrule
% Hillary Clinton & RT $@$user: If Hillary IS elected, I have my transition team together to smoothly move me from a racist to a sexist. & -1 & -3 & 1 \\
% \hline
\end{tabular}}
\caption{Examples of sentence--target pairs with diverging aggregated scores for trust (T) and sociability (S) on one hand, and competence (C) on the other. The contrast illustrates cases where a single sentence expresses both moral or social condemnation (i.e., low T and S scores) and an acknowledgment of ability (i.e., medium to high C scores).}
\vspace*{-3mm}
\label{tab:div-comp}
\end{table}

The strength of the correlation scores presented in \autoref{tab:spearman-correlations} is also observed in the distributions of co-occurring scores shown in Appendix \ref{appendix:corrs}. Across all grids, neutral judgments co-occur most frequently, indicating in particular that neutral competence is typically paired with neutral trust and sociability. The trust--sociability grid exhibits the most compact diagonal structure, with very few opposite-sign judgments, confirming that these traits are closely intertwined in practice. By contrast, the trust--competence and sociability--competence grids display greater dispersion, which helps explain their weaker correlations and allows for cases in which competence coexists with distrust or unsociability. These less compact diagonals also shed light on instances of divergent warmth and competence judgments discussed in \autoref{tab:div-comp}.

\begin{table}[!t]
\centering
\small
\begin{tabular}{l c}
\toprule
\textbf{Pair of Dimensions} & \textbf{Spearman $\rho$} \\
\midrule
Trust and Sociability & 0.79 \\
Sociability and Competence & 0.68 \\
Trust and Competence & 0.67 \\
\bottomrule
\end{tabular}
\caption{Spearman correlation coefficients between the annotated social dimensions. All correlations are positive and relatively strong.}
\vspace*{-3mm}
\label{tab:spearman-correlations-pairs}
\end{table}

\section{Conclusion}
We introduced W\&C-Sent, the first sentence-level dataset annotated for social perception in text, with over 1,600 instances covering seven target individuals and social groups. Each instance is annotated by fluent English speakers for perceptions of trust, sociability, and competence. Our experiments demonstrate the utility of our dataset and show that models struggle to capture subtle social cues.

We make W\&C-Sent publicly available to support further research on how language conveys social perceptions, enabling applications such as text analytics, bias and stereotype evaluation, and socially aware NLP systems. W\&C-Sent also serves as a benchmark for assessing whether NLP models and LLMs accurately capture nuanced social traits.

\section*{Limitations}
Our dataset focuses on English and relies on human annotations. While we collected responses from a substantial number of annotators ($n=210$), most come from English-speaking regions commonly referred to as the ``Global North''. Social perceptions are inherently culture-specific and subjective, even within the same cultural context. Hence, our annotations reflect the annotators' lived experiences and intuitions regarding warmth and competence. We do not claim that they capture all possible perceptions; rather, they provide a well-defined, high-quality baseline for English-speaking populations. By releasing W\&C-Sent along with individual annotations, we provide opportunities for future work to explore how annotators perceive socio-linguistic cues. Systematic investigations of scale design could inform frameworks for nuanced social perception tasks.
\newline
W\&C-Sent mostly leverages the publicly available SemEval-2016 Stance dataset to identify sentences suitable for assessing warmth and competence. We acknowledge that the dataset primarily reflects populations with internet access and the technological means to express opinions online, which may not cover all socioeconomic contexts. Although we limited ourselves to targets present in instances of the SemEval-2016 dataset, our methodology is generalizable and can be applied to build larger resources with additional targets and scenarios. We encourage NLP researchers to extend W\&C-Sent to new targets and languages and make our guidelines and resources publicly available to facilitate this.
\newline
Finally, future research could build on our experiments by evaluating the performance of additional LLMs on trust, sociability, and competence. Our study was limited to a subset of models due to space and cost considerations.

\section*{Ethical Considerations}

Our study was conducted with approval from our Institutional Review Board (IRB) to ensure adherence to ethical research standards. All annotators were compensated fairly, at rates between 16 and 22 USD per hour---well above the minimum wage---and were provided with clear guidelines and protections. Annotators were recruited via Prolific, remained anonymous, and were informed that they could withdraw from the study at any time, receiving warnings and instructions as specified in the guidelines.
\newline
We acknowledge that social perception annotations are inherently subjective and culture-specific, as discussed in our limitations. Therefore, interpreting or generalizing results within and beyond the English-speaking populations represented in our dataset must be conducted with caution. We encourage responsible use and careful contextualization of findings in any downstream applications, and ethical reflection is strongly recommended before using our dataset. Use of the data for commercial purposes or by state actors in high-risk applications is strictly prohibited unless explicitly approved by the dataset creators. Systems developed using our datasets may not be reliable at the individual instance level and are sensitive to domain shifts. They should not be used to make critical decisions about individuals, such as in health-related applications, without appropriate expert oversight. For a comprehensive discussion on these issues, refer to \cite{mohammad-2022-ethics-sheet,mohammad2023best}.
\newline
For tasks such as spelling and grammar correction, we used privacy-protected AI assistants to ensure confidentiality.

\section*{Acknowledgments}
We thank Jan Philip Wahle for granting us early access to the ABCDE dataset. 
\newline
Thanks to Zara Siddique and Jan Philip Wahle for helpful discussions during the early stages of this project.
\newpage
\bibliography{anthology,custom}

\begin{thebibliography}{35}
\providecommand{\natexlab}[1]{#1}

\bibitem[{Abele et~al.(2016)Abele, Hauke, Peters, Louvet, Szymkow, and Duan}]{agency-communion}
Andrea~E. Abele, Nicole Hauke, Kim Peters, Eva Louvet, Aleksandra Szymkow, and Yanping Duan. 2016.
\newblock \href {https://doi.org/10.3389/fpsyg.2016.01810} {{F}acets of the {F}undamental {C}ontent {D}imensions: {A}gency with {C}ompetence and {A}ssertiveness—{C}ommunion with {W}armth and {M}orality}.
\newblock \emph{{F}rontiers in {P}sychology}, Volume 7 - 2016.

\bibitem[{Blodgett et~al.(2020)Blodgett, Barocas, Daum{\'e}~III, and Wallach}]{blodgett-etal-2020-language}
Su~Lin Blodgett, Solon Barocas, Hal Daum{\'e}~III, and Hanna Wallach. 2020.
\newblock \href {https://doi.org/10.18653/v1/2020.acl-main.485} {Language (technology) is power: A critical survey of {``}bias{''} in {NLP}}.
\newblock In \emph{Proceedings of the 58th Annual Meeting of the Association for Computational Linguistics}, pages 5454--5476, Online. Association for Computational Linguistics.

\bibitem[{Cheng et~al.(2023)Cheng, Durmus, and Jurafsky}]{cheng-etal-2023-marked}
Myra Cheng, Esin Durmus, and Dan Jurafsky. 2023.
\newblock \href {https://doi.org/10.18653/v1/2023.acl-long.84} {Marked personas: Using natural language prompts to measure stereotypes in language models}.
\newblock In \emph{Proceedings of the 61st Annual Meeting of the Association for Computational Linguistics (Volume 1: Long Papers)}, pages 1504--1532, Toronto, Canada. Association for Computational Linguistics.

\bibitem[{Cronbach(1951)}]{Cronbach1951CoefficientAA}
Lee~Joseph Cronbach. 1951.
\newblock \href {https://api.semanticscholar.org/CorpusID:13820448} {Coefficient alpha and the internal structure of tests}.
\newblock \emph{Psychometrika}, 16:297--334.

\bibitem[{Cuddy et~al.(2007)Cuddy, Fiske, and Glick}]{Cuddy2007TheBM}
Amy J.~C. Cuddy, Susan~T. Fiske, and Peter Glick. 2007.
\newblock \href {https://api.semanticscholar.org/CorpusID:16399286} {The bias map: behaviors from intergroup affect and stereotypes.}
\newblock \emph{Journal of personality and social psychology}, 92 4:631--48.

\bibitem[{Fiske(2018)}]{fiske-endure-2018}
Susan~T. Fiske. 2018.
\newblock \href {https://doi.org/10.1177/0963721417738825} {{S}tereotype {C}ontent: {W}armth and {C}ompetence {E}ndure}.
\newblock \emph{{C}urrent {D}irections in {P}sychological {S}cience}, 27(2):67--73.
\newblock PMID: 29755213.

\bibitem[{Fiske et~al.(2002)Fiske, Cuddy, Glick, and Xu}]{Fiske2002}
Susan~T. Fiske, Amy J.~C. Cuddy, Peter Glick, and Jun Xu. 2002.
\newblock \href {https://doi.org/10.1037/pspa0000163} {{A} model of (often mixed) stereotype content: competence and warmth respectively follow from perceived status and competition}.
\newblock \emph{{J}ournal of {P}ersonality and {S}ocial {P}sychology}, 82(6):878--902.
\newblock Erratum in: J Pers Soc Psychol. 2024 Mar;126(3):412.

\bibitem[{Fraser et~al.(2024)Fraser, Kiritchenko, and Nejadgholi}]{fraser-etal-2024-stereotype}
Kathleen Fraser, Svetlana Kiritchenko, and Isar Nejadgholi. 2024.
\newblock \href {https://doi.org/10.18653/v1/2024.starsem-1.2} {How does stereotype content differ across data sources?}
\newblock In \emph{Proceedings of the 13th Joint Conference on Lexical and Computational Semantics (*SEM 2024)}, pages 18--34, Mexico City, Mexico. Association for Computational Linguistics.

\bibitem[{{Gemma Team}(2025)}]{gemma_2025}
{Gemma Team}. 2025.
\newblock \href {https://goo.gle/Gemma3Report} {{G}emma 3}.

\bibitem[{Kiritchenko and Mohammad(2018)}]{kiritchenko-mohammad-2018-examining}
Svetlana Kiritchenko and Saif Mohammad. 2018.
\newblock \href {https://doi.org/10.18653/v1/S18-2005} {Examining gender and race bias in two hundred sentiment analysis systems}.
\newblock In \emph{Proceedings of the Seventh Joint Conference on Lexical and Computational Semantics}, pages 43--53, New Orleans, Louisiana. Association for Computational Linguistics.

\bibitem[{Koch et~al.(2024)Koch, Smith, Fiske, Abele, Ellemers, and Yzerbyt}]{koch-morality}
Alex Koch, Austin Smith, Susan Fiske, Andrea Abele, Naomi Ellemers, and Vincent Yzerbyt. 2024.
\newblock \href {https://doi.org/10.31234/osf.io/8ejbn} {{V}alidating a brief measure of four facets of social evaluation}.
\newblock \emph{{B}ehavior {R}esearch {M}ethods}.

\bibitem[{Kuder and Richardson(1937)}]{Kuder1937TheTO}
G.~Frederic Kuder and Marion~W. Richardson. 1937.
\newblock \href {https://api.semanticscholar.org/CorpusID:121460737} {The theory of the estimation of test reliability}.
\newblock \emph{Psychometrika}, 2:151--160.

\bibitem[{Liu et~al.(2023)Liu, Lin, Hewitt, Paranjape, Bevilacqua, Petroni, and Liang}]{lost-in-the-middle}
Nelson~F. Liu, Kevin Lin, John Hewitt, Ashwin Paranjape, Michele Bevilacqua, Fabio Petroni, and Percy Liang. 2023.
\newblock \href {https://arxiv.org/abs/2307.03172} {{L}ost in the {M}iddle: {H}ow {L}anguage {M}odels {U}se {L}ong {C}ontexts}.
\newblock \emph{Preprint}, arXiv:2307.03172.

\bibitem[{Mohammad(2018)}]{vad-2018}
Saif Mohammad. 2018.
\newblock \href {https://doi.org/10.18653/v1/P18-1017} {{O}btaining {R}eliable {H}uman {R}atings of {V}alence, {A}rousal, and {D}ominance for 20,000 {E}nglish {W}ords}.
\newblock In \emph{{P}roceedings of the 56th {A}nnual {M}eeting of the {A}ssociation for {C}omputational {L}inguistics ({V}olume 1: {L}ong {P}apers)}, pages 174--184, Melbourne, Australia. {A}ssociation for {C}omputational {L}inguistics.

\bibitem[{Mohammad(2023)}]{mohammad2023best}
Saif Mohammad. 2023.
\newblock Best practices in the creation and use of emotion lexicons.
\newblock In \emph{Findings of the Association for Computational Linguistics: EACL 2023}, pages 1825--1836.

\bibitem[{Mohammad(2022)}]{mohammad-2022-ethics-sheet}
Saif~M. Mohammad. 2022.
\newblock \href {https://doi.org/10.1162/coli_a_00433} {Ethics sheet for automatic emotion recognition and sentiment analysis}.
\newblock \emph{Computational Linguistics}, 48(2):239--278.

\bibitem[{Mohammad(2025{\natexlab{a}})}]{nrcvadlexiconv2}
Saif~M. Mohammad. 2025{\natexlab{a}}.
\newblock \href {https://arxiv.org/abs/2503.23547} {{N}{R}{C} {V}{A}{D} {L}exicon v2: {N}orms for {V}alence, {A}rousal, and {D}ominance for over 55k {E}nglish {T}erms}.
\newblock \emph{Preprint}, arXiv:2503.23547.

\bibitem[{Mohammad(2025{\natexlab{b}})}]{wordsofwarmth}
Saif~M. Mohammad. 2025{\natexlab{b}}.
\newblock \href {https://arxiv.org/abs/2506.03993} {{W}ords of {W}armth: {T}rust and {S}ociability {N}orms for over 26k {E}nglish {W}ords}.
\newblock \emph{Preprint}, arXiv:2506.03993.

\bibitem[{Mohammad et~al.(2016)Mohammad, Kiritchenko, Sobhani, Zhu, and Cherry}]{StanceSemEval2016}
Saif~M. Mohammad, Svetlana Kiritchenko, Parinaz Sobhani, Xiaodan Zhu, and Colin Cherry. 2016.
\newblock {S}emeval-2016 {T}ask 6: {D}etecting {S}tance in {T}weets.
\newblock In \emph{{P}roceedings of the {I}nternational {W}orkshop on {S}emantic {E}valuation}, SemEval '16, San Diego, California.

\bibitem[{Nadeem et~al.(2021)Nadeem, Bethke, and Reddy}]{stereoset}
Moin Nadeem, Anna Bethke, and Siva Reddy. 2021.
\newblock \href {https://doi.org/10.18653/v1/2021.acl-long.416} {{S}tereo{S}et: {M}easuring stereotypical bias in pretrained language models}.
\newblock In \emph{{P}roceedings of the 59th {A}nnual {M}eeting of the {A}ssociation for {C}omputational {L}inguistics and the 11th {I}nternational {J}oint {C}onference on {N}atural {L}anguage {P}rocessing ({V}olume 1: {L}ong {P}apers)}, pages 5356--5371, Online. {A}ssociation for {C}omputational {L}inguistics.

\bibitem[{Nguyen et~al.(2020)Nguyen, Vu, and Nguyen}]{bertweet}
Dat~Quoc Nguyen, Thanh Vu, and Anh~Tuan Nguyen. 2020.
\newblock \href {https://arxiv.org/abs/2005.10200} {{{B}{E}{R}{T}}weet: {A} pre-trained language model for {E}nglish {T}weets}.
\newblock \emph{Preprint}, arXiv:2005.10200.

\bibitem[{Nicolas et~al.(2021)Nicolas, Bai, and Fiske}]{nicolas-dictionary}
Gandalf Nicolas, Xuechunzi Bai, and Susan~T. Fiske. 2021.
\newblock \href {https://doi.org/10.1002/ejsp.2724} {{C}omprehensive stereotype content dictionaries using a semi-automated method}.
\newblock \emph{{E}uropean {J}ournal of {S}ocial {P}sychology}, 51(1):178--196.

\bibitem[{OpenAI(2023)}]{gpt4}
OpenAI. 2023.
\newblock \href {https://doi.org/10.48550/ARXIV.2303.08774} {{GPT-4} technical report}.
\newblock \emph{CoRR}, abs/2303.08774.

\bibitem[{Ousidhoum et~al.(2021)Ousidhoum, Zhao, Fang, Song, and Yeung}]{ousidhoum-etal-2021-probing}
Nedjma Ousidhoum, Xinran Zhao, Tianqing Fang, Yangqiu Song, and Dit-Yan Yeung. 2021.
\newblock \href {https://doi.org/10.18653/v1/2021.acl-long.329} {Probing toxic content in large pre-trained language models}.
\newblock In \emph{Proceedings of the 59th Annual Meeting of the Association for Computational Linguistics and the 11th International Joint Conference on Natural Language Processing (Volume 1: Long Papers)}, pages 4262--4274, Online. Association for Computational Linguistics.

\bibitem[{Plaza-del Arco et~al.(2024{\natexlab{a}})Plaza-del Arco, Cercas~Curry, Curry, Abercrombie, and Hovy}]{plaza-del-arco-etal-2024-angry}
Flor~Miriam Plaza-del Arco, Amanda Cercas~Curry, Alba Curry, Gavin Abercrombie, and Dirk Hovy. 2024{\natexlab{a}}.
\newblock \href {https://doi.org/10.18653/v1/2024.acl-long.415} {Angry men, sad women: Large language models reflect gendered stereotypes in emotion attribution}.
\newblock In \emph{Proceedings of the 62nd Annual Meeting of the Association for Computational Linguistics (Volume 1: Long Papers)}, pages 7682--7696, Bangkok, Thailand. Association for Computational Linguistics.

\bibitem[{Plaza-del Arco et~al.(2024{\natexlab{b}})Plaza-del Arco, Curry, Paoli, Cercas~Curry, and Hovy}]{plaza-del-arco-etal-2024-divine}
Flor~Miriam Plaza-del Arco, Amanda~Cercas Curry, Susanna Paoli, Alba Cercas~Curry, and Dirk Hovy. 2024{\natexlab{b}}.
\newblock \href {https://doi.org/10.18653/v1/2024.findings-emnlp.251} {Divine {LL}a{MA}s: Bias, stereotypes, stigmatization, and emotion representation of religion in large language models}.
\newblock In \emph{Findings of the Association for Computational Linguistics: EMNLP 2024}, pages 4346--4366, Miami, Florida, USA. Association for Computational Linguistics.

\bibitem[{{Qwen Team}(2024)}]{qwen2.5}
{Qwen Team}. 2024.
\newblock \href {https://qwenlm.github.io/blog/qwen2.5/} {{Q}wen2.5: {A} {P}arty of {F}oundation {M}odels}.

\bibitem[{Siddique et~al.(2024)Siddique, Turner, and Espinosa-Anke}]{siddique-etal-2024-better}
Zara Siddique, Liam Turner, and Luis Espinosa-Anke. 2024.
\newblock \href {https://doi.org/10.18653/v1/2024.emnlp-main.1035} {Who is better at math, jenny or jingzhen? uncovering stereotypes in large language models}.
\newblock In \emph{Proceedings of the 2024 Conference on Empirical Methods in Natural Language Processing}, pages 18601--18619, Miami, Florida, USA. Association for Computational Linguistics.

\bibitem[{Singh et~al.(2025)Singh, Fry, Perelman, Tart, Ganesh, El-Kishky, McLaughlin, Low, Ostrow, Ananthram et~al.}]{singh2025openai}
Aaditya Singh, Adam Fry, Adam Perelman, Adam Tart, Adi Ganesh, Ahmed El-Kishky, Aidan McLaughlin, Aiden Low, AJ~Ostrow, Akhila Ananthram, et~al. 2025.
\newblock Openai gpt-5 system card.
\newblock \emph{arXiv preprint arXiv:2601.03267}.

\bibitem[{Szab{\'o}(2024)}]{Szabo2024Compositionality}
Zolt{\'a}n~Gendler Szab{\'o}. 2024.
\newblock \href {https://plato.stanford.edu/archives/sum2024/entries/compositionality/} {Compositionality}.
\newblock In Edward~N. Zalta and Uri Nodelman, editors, \emph{The {Stanford} Encyclopedia of Philosophy}, summer 2024 edition. Metaphysics Research Lab, Stanford University.

\bibitem[{Team(2025)}]{qwen3technicalreport}
Qwen Team. 2025.
\newblock \href {https://arxiv.org/abs/2505.09388} {Qwen3 technical report}.
\newblock \emph{Preprint}, arXiv:2505.09388.

\bibitem[{Teodorescu and Mohammad(2023)}]{teodorescu-mohammad-2023-evaluating}
Daniela Teodorescu and Saif Mohammad. 2023.
\newblock \href {https://doi.org/10.18653/v1/2023.findings-emnlp.271} {Evaluating emotion arcs across languages: Bridging the global divide in sentiment analysis}.
\newblock In \emph{Findings of the Association for Computational Linguistics: EMNLP 2023}, pages 4124--4137, Singapore. Association for Computational Linguistics.

\bibitem[{Wahle et~al.(2026)Wahle, Vishnubhotla, Gipp, and Mohammad}]{wahle-etal-2026-abcde}
Jan~Philip Wahle, Krishnapriya Vishnubhotla, Bela Gipp, and Saif~M. Mohammad. 2026.
\newblock Affect, body, cognition, demographics, and emotion: The abcde of text features for computational affective science.
\newblock In \emph{Proceedings of the 1st Workshop on Computational Affective Science (CAS 2026)}, Palma de Mallorca, Spain. European Language Resources Association (ELRA).

\bibitem[{Weir(2005)}]{Weir2005QuantifyingTR}
Joseph~P. Weir. 2005.
\newblock \href {https://api.semanticscholar.org/CorpusID:1420444} {Quantifying test-retest reliability using the intraclass correlation coefficient and the sem.}
\newblock \emph{Journal of strength and conditioning research}, 19 1:231--40.

\bibitem[{Wolf et~al.(2019)Wolf, Debut, Sanh, Chaumond, Delangue, Moi, Cistac, Rault, Louf, Funtowicz et~al.}]{huggingface}
Thomas Wolf, Lysandre Debut, Victor Sanh, Julien Chaumond, Clement Delangue, Anthony Moi, Pierric Cistac, Timoth{\'e}e Rault, R{\'e}mi Louf, Morgan Funtowicz, et~al. 2019.
\newblock {H}ugging{F}ace's {T}ransformers: {S}tate-of-the-art {N}atural {L}anguage {P}rocessing.
\newblock \emph{ar{X}iv preprint ar{X}iv:1910.03771}.

\end{thebibliography}

\newpage
\appendix
\section*{Appendix}
\section{Selection from ABCDE Dataset}

We extracted 108 sentences about Donald Trump from the ABCDE dataset, representing 26.34\% of all sentences for that target and 71.52\% of the sentences extracted from ABCDE.

\begin{table}[ht]
% \centering
\small
\resizebox{\columnwidth}{!}{%
\begin{tabular}{|>{\centering\arraybackslash}m{2.5cm}|m{5cm}|}\hline
\textbf{Target} & \multicolumn{1}{c|}{\textbf{Keywords}}  \\
\hline
Women & girl, girls, woman, women, feminism, feminist \\ \hline
Religious People and nonreligious People & religious, religion, God, nonreligious, atheism, atheist, Christian, Christianity, Christians, Muslim, Islam, Muslims, Bible \\ \hline
Hillary Clinton & Hillary, Clinton, Hillary Clinton, HC, Benghazi \\ \hline
Donald Trump & Donald, Trump, Donald Trump \\ \hline
Barack Obama & Obama, Barack, Barack Obama, Obamacare \\ \hline
Environmentalists & environment, environmental, environmentalist, environmentalism, global warming, climate change \\ \hline
\end{tabular}}
\caption{Keywords using which the sentences from ABCDE were selected.}
\label{tab:abcde-keywords}

\end{table}

\label{appendix:abcde-selection}

\section{Annotator Selection in Prolific}

The filters described below were put in place to ensure that participants meet the language requirements and demonstrate strong qualifications and proven reliability. 
%\newline

\paragraph{1. \quad Countries.}

The first two screener sets selected were the ``current country of residence" and the ``country of birth". Since the success of the task hinges on fluency in English, the countries selected in this set were those that mainly speak English, so the following countries were selected in both sets: Antigua and Barbuda, Australia, Barbados, Belize, Canada, Ireland, the United Kingdom, the United States, and New Zealand.%\newline

\paragraph{2. \quad Languages.}

Prolific offers three screener sets related to the languages that annotators speak. Those are ``first language", ``primary language", and ``fluent languages". Those were all set to ``English".

Prolific displays the number of eligible participants after each screener set is applied, and the number decreases with every additional filter. 
% This indicates that the participants in this study satisfied all three requirements simultaneously, rather than qualifying through only one or a combination of them.%\newline

\paragraph{3. \quad Education.}

For most screener sets regarding ``highest education level completed", eligible participants were limited to those with at least a technical or community college degree, ranging up through undergraduate, graduate, and doctoral qualifications.

\paragraph{4. \quad Approval Rate and Participation.}

The approval rate and previous participation criteria were also important in shaping the pool of annotators. By requiring a 99-100\% approval rate, I attempted to minimize the risk of low-quality or careless responses, admitting only participants with an almost impeccable record who completed tasks to researchers' satisfaction. The additional restriction of having over 500 previously approved submissions (later increased to 2000) further ensured that participants were not only reliable but also highly experienced with research tasks on Prolific.

\label{appendix:ann-sel}

% \newpage
% -
% \newpage
\section{Data Distribution}

\begin{table}[hbt!]
\centering
% \small
\renewcommand{\arraystretch}{1.2}
\resizebox{\columnwidth}{!}{%
\begin{tabular}{|>{\centering\arraybackslash}m{2.2cm}|>{\centering\arraybackslash}m{1.2cm}|>{\centering\arraybackslash}m{1.2cm}|>{\centering\arraybackslash}m{1.2cm}|>{\centering\arraybackslash}m{1.2cm}|>{\centering\arraybackslash}m{1.5cm}|>{\centering\arraybackslash}m{1.5cm}|}
\hline
 & Hillary Clinton & Donald Trump & Feminist Movement & Atheism & Legalization of Abortion & Climate Change \\ \hline
Hillary Clinton      & 594 & 13  & 1   & 0  & 2   & 0  \\ \hline
Donald Trump         & 39  & 260 & 0   & 1  & 0   & 1  \\ \hline
Women                & 25  & 4   & 151 & 5  & 106 & 1  \\ \hline
Barack Obama         & 48  & 38  & 0   & 2  & 9   & 12 \\ \hline
Religious People     & 2   & 3   & 20  & 89 & 1   & 3  \\ \hline
Environm-entalists   & 2   & 0   & 0   & 0  & 0   & 35 \\ \hline
Nonreligious People  & 0   & 1   & 1   & 33 & 0   & 0  \\ \hline
\end{tabular}}
\caption{Distribution of sentences from the original five targets (columns) across the expanded set of seven targets in \textit{W\&C-Sent} (rows).}
\label{tab:per-target-breakdown}
\end{table}

% \begin{table*}[hbt!]
% \centering
% \small
% \resizebox{\columnwidth}{!}{%
% \renewcommand{\arraystretch}{1.2} 
% \begin{tabular}{|>{\centering\arraybackslash}m{2cm}|>{\centering\arraybackslash}m{1.5cm}|>{\centering\arraybackslash}m{1.5cm}|>{\centering\arraybackslash}m{1.5cm}|>{\centering\arraybackslash}m{1.5cm}|>{\centering\arraybackslash}m{2cm}|>{\centering\arraybackslash}m{2cm}|>{\centering\arraybackslash}m{2cm}|}
% \hline
%  & Hillary Clinton    & Donald Trump    & Women & Barack Obama & Religious People & Environm-entalists & Nonreligious People\\ \hline
% Hillary Clinton             & 594   & 39    & 25    & 48    & 2     & 2     & 0 \\ \hline
% Donald Trump                & 13    & 260   & 4     & 38    & 3     & 0    & 1\\ \hline
% Feminist Movement           & 1     & 0     & 151   & 0     & 20    & 0    &  1\\ \hline
% Atheism                     & 0     & 1     & 5     & 2     & 89    & 0    & 33\\ \hline
% Legalization of Abortion    & 2     & 0     & 106   & 9     & 1     & 0    & 0\\ \hline
% Climate Change              & 0     & 1     & 1     & 12    & 3     & 35    & 0\\ \hline
% \end{tabular}}
% \caption{Distribution of sentences from the original five targets (rows) across the expanded set of seven targets in \textit{W\&C-Sent} (columns).}
% \label{tab:per-target-breakdown}
% \end{table*}

\begin{table}[hbt!]
\centering
% \small
\resizebox{\columnwidth}{!}{
\begin{tabular}{|>{\centering\arraybackslash}m{1cm}|>{\centering\arraybackslash}m{3cm}|m{4cm}|c|} \hline
\textbf{ID} & \textbf{Original Target} & \hfil \textbf{Sample sentence} & \textbf{Decision}\\ \hline
288 & Atheism & I will dwell in a peaceful habitation, in secure dwellings, and in quiet resting places -Isa. 32:18 & Not selected\\ \hline
11182 & Legalization of Abortion & Living in a pub isnt all that good when your friends turn into alcoholics \#noonewillunderstand & Not selected \\ \hline
1463 & Feminist Movement & We are 51\% of the population and only 17\% of Congress. The \#WarOnWomen is absolutely a real thing. Wake up, America. & Selected \\\hline
1710 & Hillary Clinton & Hillary is killing it so far on the trail. She's finally showing her personal side and I think it will benefit her profoundly. & Selected \\ \hline
\end{tabular}}
\caption{Examples of sentences from the SemEval-2016 stance dataset that were selected and not selected into W\&C-Sent, with their original targets and IDs.}
\label{tab:samples-selected}
\end{table}

\newpage

\section{The Complete English Annotation Guidelines}

\label{appendix:english-guidelines}

{\large \textbf{Preliminary instructions}}

\begin{enumerate}
  \item Attempt these questions only if you are fluent in English. 
  \item Your responses are confidential.
  \item There is a degree of subjectivity in this task. Let your instinct guide you; don’t overthink it. 
  \item Consider the entire meaning of the sentence before attempting to give the relevant scores.
  \item Your views regarding any of the entities or topics in the texts (such as political parties, individuals, social groups) should \underline{not} affect your scores. 
  \item To ensure fairness and the validity of our scientific findings, some questions (typically unambiguous ones!) have predetermined answer ranges. While occasional deviations are acceptable given the subjectivity of this task, contributions may be rejected if a considerable number of these questions are answered incorrectly. Reading the guidelines below is therefore essential for a successful participation and compensation. This measure will ensure honest participation is compensated fairly. \newline

\end{enumerate}

\noindent {\large Task definition and theoretical background}

\begin{itemize}
    \item Social psychology research has shown that individuals rapidly and subconsciously evaluate others, groups, and even themselves along the dimensions of \textbf{warmth (W)} and \textbf{competence (C)}.

    \item Recently, psychologists have modeled warmth through two dimensions: \textbf{trust (T)} and \textbf{sociability (S)}.\footnote{This part was excluded from the competence-specific guidelines as it is irrelevant to the task.}

    \item This task will aim to assess the degrees of \textbf{ \underline{trust}}, \textbf{ \underline{sociability}}, and \textbf{ \underline{competence}} towards a \textbf{ \underline{specific target} within a sentence}. %\newpage
    \item \underline{\textbf{Trust}}:\footnotemark
    \begin{itemize}
        \item[◦] The focus in this dimension is on the personal / moral aspect of the target.
        \item[◦] \textit{High trust} can be defined as morality, kindness, sincerity, trustworthiness, and honesty. 
        \item[◦] \underline{Words associated with high trust}: charity, mother, compliment, affectionate.
        \item[◦] \textit{Low trust} can be defined as immorality, insincerity, dishonesty, untrustworthiness, dubiousness, and maliciousness.
        \item[◦]  \underline{Words associated with low trust}: discredit, bribe, espionage, disinformation, disloyal.
    \end{itemize}

    \item \underline{\textbf{Sociability}}:\footnotemark[\value{footnote}]
    \begin{itemize}
        \item[◦] The focus in this dimension is on the social aspect of the target and its relational impact on others or society as a whole.
    
        \item[◦] \textit{High sociability} can be defined as friendliness, sociableness, generosity, and helpfulness. 
    
        \item[◦] \underline{Words associated with high sociability}: helpful, intimate, laugh, celebration, reliant, entertain, social club, bestie. 
    
        \item[◦] \textit{Low sociability} can be defined as antisocial behavior, lack of generosity, inconsiderateness, indifference, and unhelpfulness.
    
        \item[◦] \underline{Words associated with low sociability}: ingrate, abduct, selfish, theft, egomaniacal, pervert.

    \end{itemize}

    \item \underline{\textbf{Competence}}\footnotemark[\value{footnote}]:

    \begin{itemize}
        \item[◦] \textit{Competence} can be defined as ability, power, dominance, being in control, importance, having influence, and assertiveness. 
    
        \item[◦] \underline{Words associated with high competence}: hitman, heroical, entrepreneurship, strategies, superman, viper, impunity.
    
        \item[◦] \textit{Incompetence} can be defined as submissiveness, not being in control of a situation, being controlled or guided by outside factors, or weakness.
    
        \item[◦] \underline{Words associated with low competence}: bootlicker, talentless, crash landing, bedridden, underfed. 
        % \newpage
    
    \end{itemize}
\end{itemize}

% \newpage
\noindent {\large The task}

\begin{itemize}
    \item You will be given a text snippet and a target (group, entity, or individual). Below, you will be able to assign scores. Please rate the apparent levels of trust, sociability, or competence\footnotemark[\value{footnote}] that the sentence's author seems to express towards the specified target.

    \item \underline{The labels}:\footnotemark[\value{footnote}]

    \begin{itemize}
        \item[◦] -3	 high    distrust / unsociability / incompetence
        \item[◦] -2	 moderate     distrust / unsociability / incompetence
        \item[◦] -1	 slight    distrust / unsociability / incompetence
        \item[◦]  0	 neutral, not applicable, not expressed, etc.
        \item[◦] +1	 slight    trust / sociability / competence
        \item[◦] +2  moderate    trust / sociability / competence
        \item[◦] +3  high    trust / sociability / competence
    
    \end{itemize}
\end{itemize}

\noindent For the sentences in this task, the targets will be one of the following: 

\begin{enumerate}
    \item \underline{\textit{Religious people}}: this target will be listed along with sentences that are related to religions and target those who believe in God or practice (any) religion.
    \item \underline{\textit{Nonreligious people}}: this target will be listed along with sentences that are related to atheism and target those who are not religious (i.e., atheists/agnostics).
    \item \underline{\textit{Women}}: this target will be listed along with sentences that are related to women or girls and/or interconnected topics: sexism, feminism, misogyny, bias, and female representation. 
    \item \underline{\textit{Environmentalists}}: this target will be listed along with sentences that are related to environment and climate change activists.
    \item \underline{\textit{Hillary Clinton}}: this target will be listed along with sentences that are related to the 2016 US presidential candidate and former Secretary of State Hillary Clinton.
    \item \underline{\textit{Donald Trump}}: this target will be listed along with sentences that are related to the 2016 US presidential candidate and current US president Donald Trump.
    \item \underline{\textit{Barack Obama}}: this target will be listed along with sentences that are related to the former US president Barack Obama.\newline

\end{enumerate}

\noindent In order to assess trust, sociability, and competence, \underline{try to answer the following questions}\footnotemark[\value{footnote}]:
\footnotetext{Only the relevant dimension was included in each dimension-specific guidelines..}

\begin{enumerate}
    \item What is the \underline{degree of trust} \textbf{towards this target} that \textbf{the author} of the text seems to express? Does \textbf{the author} seem to perceive the target as trustworthy or untrustworthy / moral or immoral / honest or dishonest?

    \item What is the \underline{degree of sociability} \textbf{towards this target} that \textbf{the author} of the text seems to express? Does \textbf{the author} seem to perceive the target as sociable or antisocial? Helpful or unhelpful? 

    \item What is the \underline{degree of competence} \textbf{towards this target} that \textbf{the author} seems to express? Does \textbf{the author} seem to perceive the target as in control or out of control? Active or passive? Powerful or weak?

\end{enumerate}

\noindent \underline{Notes}:
\begin{enumerate}
    \item There are select examples in the next page, accompanied by an explanation of the scores given for each example.

    \item Adhere to the literal meaning of competence, which may be ``positive" (e.g., a CEO) or ``negative" (e.g., a villain or a dictator). Both types are considered ``competence", regardless of the outcomes. Example 3 is an example of that.\footnote{This point only appeared in the competence-specific guidelines.}

    \item There are no repeated sentences in this study. All sentences were carefully chosen\footnote{This part was added after a participant in the pilot run skipped sentences they thought were repeated.}.
    
    \item Even if the speaker is explicitly expressing opinions towards X, if the target listed is Y, then we want to know the degree of trust, sociability, and competence\footnote{Only the relevant dimension was included here in each dimension-specific guidelines} towards Y only.

    \item Try to be objective. Your views regarding any of the entities or topics  in the texts (such as political parties, individuals, social groups) should not affect your scores.

    \item There is an optional free-form text field underneath each instance. You can add any comments, thoughts, or justifications you may have on the scores you gave.

    \item You will have these guidelines available to you at every stage of the task by pressing on ``See task details" on the top right.

\end{enumerate} 

% \newpage

\noindent {\large Examples}\footnote{Each dimension-specific guidelines included the relevant score and justification of that dimension only.} \newline

\noindent {\large \textbf{Example 1:}} \newline

\noindent \underline{Target}:	Women

\noindent \underline{Text}:	``My wife is the \textbf{most caring} person I've ever met ... she's the only woman in a house full of testosterone . She \textbf{never stops working} whether it's at home or being an RN . \textbf{I cant keep up} but I try . She makes me a \textbf{better person} . I'd be \textbf{lost without her} . Oh and she's smoking hot too."\newline

\noindent \underline{Trust}:	+3 (high trust). The author expresses maximum trust in women through his wife as a representative example. He portrays his wife as a \textbf{trustworthy, dependable, and caring individual} who is \textbf{essential to his well-being} (``I'd be lost without her"). The statement ``she makes me a better person" implies the author views his wife as having \textbf{a strong moral character that positively influences him}. 

\noindent \underline{Sociability}: +3 (high sociability). The text attributes maximum sociability to women through his wife. ``\textbf{Most caring person I've ever met}" shows high sociability and interpersonal warmth. He emphasizes how essential her social/emotional qualities are to their family dynamic, particularly in a ``house full of testosterone." The statement ``whether it's at home or being an RN" suggests helpfulness and generosity both personally and professionally. 

\noindent \underline{Competence}: +3 (high competence). The author says that his wife ``never stops working" in the context of her role as an RN (registered nurse) and as a mother, which suggests \textbf{multitasking, capability, and professional competence}. ``I can't keep up but I try" indicates a \textbf{highly active and energetic individual}.  \newline %\newpage

\noindent {\large \textbf{Example 2:}} \newline

\noindent \underline{Target}:	Women

\noindent \underline{Text}:	``when i was 16 i had a folder of `` \textbf{feel good songs} " and everyday i would select one and\textbf{ send it to my best friend} along with a paragraph of \textbf{how much they meant to me} and why i they should be \textbf{happy} and then i would lay in bed thinking `` no i’m not gay this is just what girls do :) "" \newline

\noindent \underline{Trust}: +3 (high trust). The author attributes \textbf{sincere, genuine intentions} to women through the described behaviors. The daily practice of \textbf{sending heartfelt messages} ``about how much they meant to me and why they should be happy" suggests honesty and sincerity in women's relationships. The author portrays women as having \textbf{good moral intentions} in their friendships by genuinely \textbf{caring about others' wellbeing}. 

\noindent \underline{Sociability}: +3 (high sociability). The text attributes maximum sociability to women. The actions described (sending feel-good songs, writing paragraphs about how much someone means to you, \textbf{wanting to make someone happy}) \textbf{represent peak social engagement and helpfulness}. The phrase ``this is just what girls do" frames these highly sociable, caring behaviors as naturally feminine traits and hence portrays women as exceptionally \textbf{generous with their emotional/social energy}.

\noindent \underline{Competence}: +2 (moderate competence). Selecting appropriate songs, writing meaningful messages, and maintaining friendships require emotional intelligence, thoughtfulness, and social skills. However, the implication that this behavior is ``just what girls do" could suggest that the author views this as \textbf{instinctual rather than skillful}, or that it's somehow slightly less significant than other types of competence.  \newpage

\noindent {\large \textbf{Example 3:}} \newline

\noindent \underline{Target}: Hillary Clinton 

\noindent \underline{Text}:	``Would you wanna be in a long term relationship with \textbf{some bitch} that \textbf{hides her emails}, \& \textbf{lies to your face}? Then \#\textbf{Dontvote}"  \newline

\noindent \underline{Trust}:	-3 (high distrust). The author explicitly portrays Clinton as someone who is \textbf{fundamentally untrustworthy} and \textbf{cannot be relied upon to tell the truth} or \textbf{be transparent} through two direct accusations: ``hides her emails" and ``lies to your face." With concealment and deception some of the \textbf{strongest markers of untrustworthiness}, the author emphasizes these as core trust violations. 

\noindent \underline{Sociability}: -2 (moderate unsociability). The derogatory term ``\textbf{bitch}" (which is demeaning towards women) and the comparison to an \textbf{undesirable romantic partner} frames Clinton as someone who would be \textbf{unpleasant to be around or interact with socially}. The rhetorical question implies she would be toxic in close social relationships. However, one can view the sociability attack is more about being unpleasant in relationships rather than being completely antisocial or unhelpful in all social contexts, hence a maximum score was not assigned.

\noindent \underline{Competence}: +2 (moderate competence). Despite the author implying that Clinton is manipulative and dishonest, the author’s phrasing doesn't suggest that she's powerless or ineffective, as the negative behaviors described (concealment and deception) require some degree of \textbf{agency and planning}. The author suggests that Clinton is being \textbf{deliberate in her (``negative") performance} and \textbf{active in her (``negative") effects}, leading to a moderately high score in the competence dimension.\newline

\noindent \underline{Notes on the suggested scores}: This example shows how your political views regarding Hillary Clinton must not influence your score. Supporters of Clinton might see this as unfair or extremely sexist, while critics might view it as more damning commentary on her performance and ability. \newpage

\noindent {\large \textbf{Example 4:}} \newline

\noindent \underline{Target}:	Women

\noindent \underline{Text}: ``My step sister \textbf{broke up} with her first boyfriend because she wanted to be independent ... \textbf{women suck} .. I'll miss you Kyle." \newline

\noindent \underline{Trust}: -1 (slight distrust). The author criticizes his sister's decision to prioritize independence and considers it hurtful to someone that they cared about, Kyle. \textbf{This suggests that the author thinks women make choices that harm others}, which is a claim towards women’s social behavior in relationships rather than their honesty or morality. 

\noindent \underline{Sociability}: -3 (high unsociability). By saying ``\textbf{women suck}", the author views women as \textbf{unpleasant, inconsiderate, or lacking in positive interpersonal qualities}. The author also frames women as \textbf{harmful to social relationships} by prioritizing their own desires over maintaining positive social connections. A maximum score was given since this is a clear negative generalization about women as a social group.

\noindent \underline{Competence}: -2 (moderate incompetence). The overall ``women suck" generalization suggests \textbf{poor judgment or decision-making by women as a social group}. This was exemplified by the author’s step-sister breaking up with someone who the author thinks she should not have. It also indicates that the author \textbf{knows better than his step-sister, and by projection, women as well}. \newline

\noindent \underline{Notes on the suggested scores}: 
\begin{enumerate}
    \item One might say that ``women suck" expresses a very negative sentiment towards women’s trustworthiness and social likeability. This might affect the scores accordingly.
    \item One might claim that the competence of women isn't really addressed since the author frames the sister's decision-making negatively rather than women as a group.
    \item Consider the fact that the gender of the author is not explicit. How might it affect your scores if the author of the post were a woman? That is up to you to decide. \newpage

\end{enumerate}

\noindent {\large \textbf{Example 5:}} \newline

\noindent \underline{Target}:	Women

\noindent \underline{Text}: ``I need feminism because the United States is one of the only countries that doesn't give paid maternity leave." \newline

\noindent \underline{Trust}:	0 (neutral). The author’s statement is focused on policy rather than character traits and doesn't make any attributions about women's trustworthiness or morality. 

\noindent \underline{Sociability}: 0 (neutral). The author’s statement is focused on policy rather than character traits. There is no commentary on women's interpersonal qualities or social behavior. 

\noindent \underline{Competence}:	 0 (neutral). The statement implies that women deserve certain rights/benefits, but it doesn't directly attribute dominance or control to women. The statement doesn't characterize women as active, powerful, passive, or weak. The author is advocating for institutional change (paid maternity leave) rather than making claims about women's power or capabilities.\newline

\noindent \underline{Note}: This is a policy statement rather than a personal attribution about women's trust, sociability, or competence. This should help you distinguish between advocacy/policy statements and personal/characteristic attributions. \newline

\noindent {\large \textbf{Example 6:}} 

\noindent \underline{Target}: Donald Trump

\noindent \underline{Text}: ``Teaching similes : ``The cats attitude was as \textbf{stubborn} as Donald Trump" \#ShitMyKidsSay" 

\noindent \underline{Trust}:	0 (neutral). The author likens Trump's stubbornness to that of a cat, which carries mild negative social judgment. However, stubbornness is not inherently about trust (morality, honesty, sincerity). Despite the playful mockery which could indicate diminished regard towards Trump, the statement doesn't make claims about Trump's reliability, honesty, or moral character.

\noindent \underline{Sociability}: -1 (slight unsociability). Stubbornness in this context suggests someone who is, like a cat, \textbf{difficult to deal with interpersonally}: not very cooperative, helpful, or considerate in social interactions. This is a \textbf{mild social criticism} that does not completely portray him as completely antisocial or unhelpful.

\noindent \underline{Competence}: -1 (slight incompetence). Comparing a political figure to a stubborn cat frames Trump as childish and unreasonable. The comparison could be seen as deliberately insulting to Trump’s \textbf{behavioral rigidity}, since a good leader is expected to be willing to change their mind based on the available evidence and opinions of experts. This shows that the author believes that Trump might \textbf{not be the best person to be the president} of the United States. 

\noindent \underline{Notes on the suggested scores}: 
\begin{enumerate}
    \item This is another example that shows that your political views regarding Donald Trump must not influence your score. Supporters of Trump might see this as unfair or even read stubbornness as positive determination, while critics might view it as more damning commentary on his interpersonal difficulties. 
    \item Other interpretations can be just as valid. One might argue that a 0 score for competence (neutral) is appropriate; generally speaking, stubbornness is a character trait that doesn't directly relate to competence or incompetence. While it can sometimes imply determination (positive for competence), in this context it's more about being inflexible or difficult rather than capable or incapable. The comparison to a cat's attitude suggests that the author is referring to behavioral rigidity on Trump’s part. The author is not making any claims about Trump's abilities, intelligence, or effectiveness. It is up to you to determine the weight given to the humorous framing vs. the underlying comparison. \newline
    
\end{enumerate}

\noindent {\large \textbf{Example 7:}} \newline

\noindent \underline{Target}:	Religious people

\noindent \underline{Text}:		``Could all those who believe in a god \textbf{please leave}. The meeting will now continue for the \textbf{grown ups only}." \newline

\noindent \underline{Trust}:	0 (neutral). The criticism is entirely focused on intellectual maturity rather than character or morality. The author is suggesting that religious people shouldn't participate in this particular discussion. On the other hand, the author doesn't make any claims about religious people's morality, honesty, sincerity, or trustworthiness. There are no accusations of deception, dishonesty, or moral failings.

\noindent \underline{Sociability}: -3 (high unsociability). The author portrays religious people as \textbf{socially incompetent} and \textbf{needing exclusion} from adult discourse (``please leave") and says their presence is incompatible with or unwanted in serious adult conversations (``grown ups only"). The order (telling an entire social group to leave) and the \textbf{dismissive language} are extremely exclusionary and socially hostile, resulting in a maximum unsociability score.

\noindent \underline{Competence}: -3 (high incompetence). Religious people are portrayed as needing to be \textbf{excluded from decision-making processes}, suggesting they lack the authority or standing or cognitive abilities to participate in important discussions. Additionally, the ``grown ups only" framing explicitly characterizes religious people as \textbf{childlike, passive, and intellectually weak} \textbf{(i.e., subordinates to ``grown ups")}. This is a direct attack on their mental capacity and maturity, which are \textbf{core competence attributes}. \newpage

\section{Distribution of Annotators' Demographic Characteristics}

\label{appendix:annotator-dist}

The annotator pool included 55.1\% identifying as male and 44.4\% as female. The majority of participants were between 25 and 44 years old (58.5\%), though the sample also included younger and older individuals up to the 65+ range. Most annotators were born in the United States (47.2\%) or the United Kingdom (38.9\%), with smaller groups from Canada (8.8\%) and other countries including Nigeria, Australia, Italy, Singapore, Ireland, Germany, Ghana, and New Zealand (5.1\%). All reported English as their primary language, and while most described themselves as monolingual (86.7\%), a minority (13.3\%) reported fluency in additional languages, such as Spanish, French, Dutch, Hindi, Urdu, and Punjabi. Further, nearly half of the annotators held a bachelor's degree (49.5\%), followed by those with a master's or community college qualification (both 18.5\%), while smaller proportions had completed high school (8.8\%) or a doctorate (4.7\%). The majority identified as White (73.6\%), with the rest distributed across Asian (8.3\%), Black (7.8\%), Mixed (6.0\%), and Other (4.1\%) categories. Table \ref{tab:demographics} below shows the complete breakdown of these attributes.

Finally, the annotators' participation among the three dimensions demonstrates broad and consistent coverage, with only a minor variation. A total of 78 contributions were received in the trust dimension, 75 in sociability, and 70 in competence.

\renewcommand{\arraystretch}{1.2}
\begin{table}[h]
\centering
\small
\resizebox{\columnwidth}{!}{
\begin{tabularx}{\linewidth}{l| *{5}{X} }
\hline
\textbf{Attribute} & \multicolumn{5}{c}{\textbf{Categories and Frequency (\%)}} \\
\hline
Gender & Male & Female & Other & \\
    & 114 (54.2) & 95 (45.2) & 1 (0.05) & \\ \hline
Age & 35-44 & 25-34 & 45-54 & Other \\
    & 66 (31.4) & 58 (27.6) & 43 (20.5) & 43 (20.5) \\  \hline
Country & US & UK & Canada & Other \\ 
        & 102 (48.5) & 78 (37.5) & 19 (9.1) & 11 (4.9) \\  \hline
Primary language & English  \\
                 & 210 (100) \\  \hline
Other languages & Monolingual & & Bilingual+ & \\ 
                 & 106 (86.2) & & 17 (13.8) & \\ \hline
Education & BA & MA & CC & HS & PhD \\
          & 102 (48.5) & 40 (19.0) & 39 (18.6) & 19 (9.1) & 10 (4.8)  \\ \hline
Ethnicity & White & Asian & Black & Other \\
          & 154 (73.3) & 18 (8.6) & 16 (7.6) & 22 (10.5) \\ \hline
Previous studies & 2000+ & 1000-1500 & 1500-2000 & 500-1000 \\
                 & 113 (53.8) & 38 (18.1) & 34 (16.2) & 25 (11.9) \\
\hline
\end{tabularx}}
\caption{Demographic breakdown of annotators. Numbers are counts with percentages in parentheses. The attribute ``Country" refers to country of birth. ``Fluent languages" refers to the languages which the annotators reported being fluent in, other than English, and not all annotators filled it. Other age ranges include 55-64 (24 annotators, or 12\%), 65+ (10 annotators, or 4.8\%), and 18-24 (9 annotators, or 4.3\%). Other ethnic groups include ``mixed" (13 annotators, or 6.2\%) and ``other" (8 annotators, or 3.8\%). In ``Education", BA = undergraduate degree, MA = graduate degree, CC = community college or technical degree, HS = high school diploma, and PhD = doctorate.}
\label{tab:demographics}
\end{table}

\section{Coarsening of Labels}

\label{appendix:coarse-procedure}

\subsection{Median-Based Coarsening}
For each sentence-target pair, all fine-grained labels were collected into a list. Then, we counted the number of negative labels ($\leq$ -1), positive labels ($\geq$ 1), and neutral labels ($=$ 0), and the majority category determined the coarse label. For example, if negative judgments were most frequent, the pair was then assigned \textit{low}. As a result, Neutral superseded in cases where neutral labels were the majority or when the counts of positive and negative labels were equal. Interestingly, there were 251 sentence-target pairs where the annotators' judgments were equally divided between \textit{high} and \textit{low}, resulting in a neutral label that reflects that lack of consensus. Almost half of those were in the competence dimension (48.6\%), while the rest was split almost equally between the sociability (26.3\%) and trust (25.1\%) dimensions.

\begin{figure}[hbt!]
   \centering
   \includegraphics[width=1\linewidth]{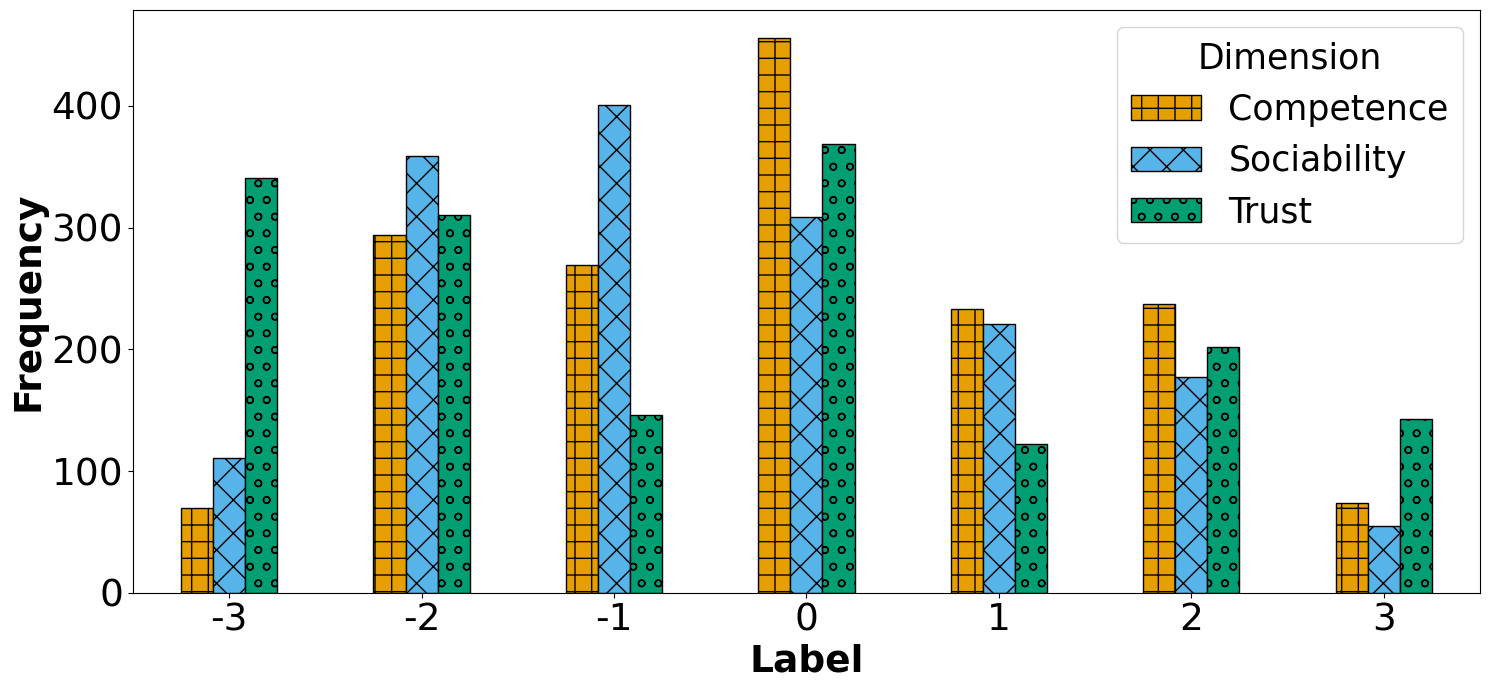}
   \caption{Distribution of discretized median-based labels, ordered from negative to positive.}
    \label{fig:median-dist}
\end{figure}

\subsection{Mean-Based Coarsening}

Following \citet{wordsofwarmth}, this method converted the average score of each sentence-target pair into seven bins (from negative to positive). This approach led to more neutral instances and a less extreme skew towards the negative side, indicating that averaging the scores smoothed the extremes of annotator judgments.

\section{Observations from the Data}

\subsection{Agreement among the Annotators}

\subsubsection{Strict Unanimity}

This refers to cases where all annotators chose the exact same fine-grained label for a sentence-target pair. 143 sentence-target pairs achieved strict unanimity, representing 3\% of the overall W\&C-Sent dataset. The most common dimension was trust (65.7\% of instances), followed by sociability (21.7\%) and competence (12.6\%).

\begin{figure}[hbt!]
    \centering
    \includegraphics[width=1\linewidth]{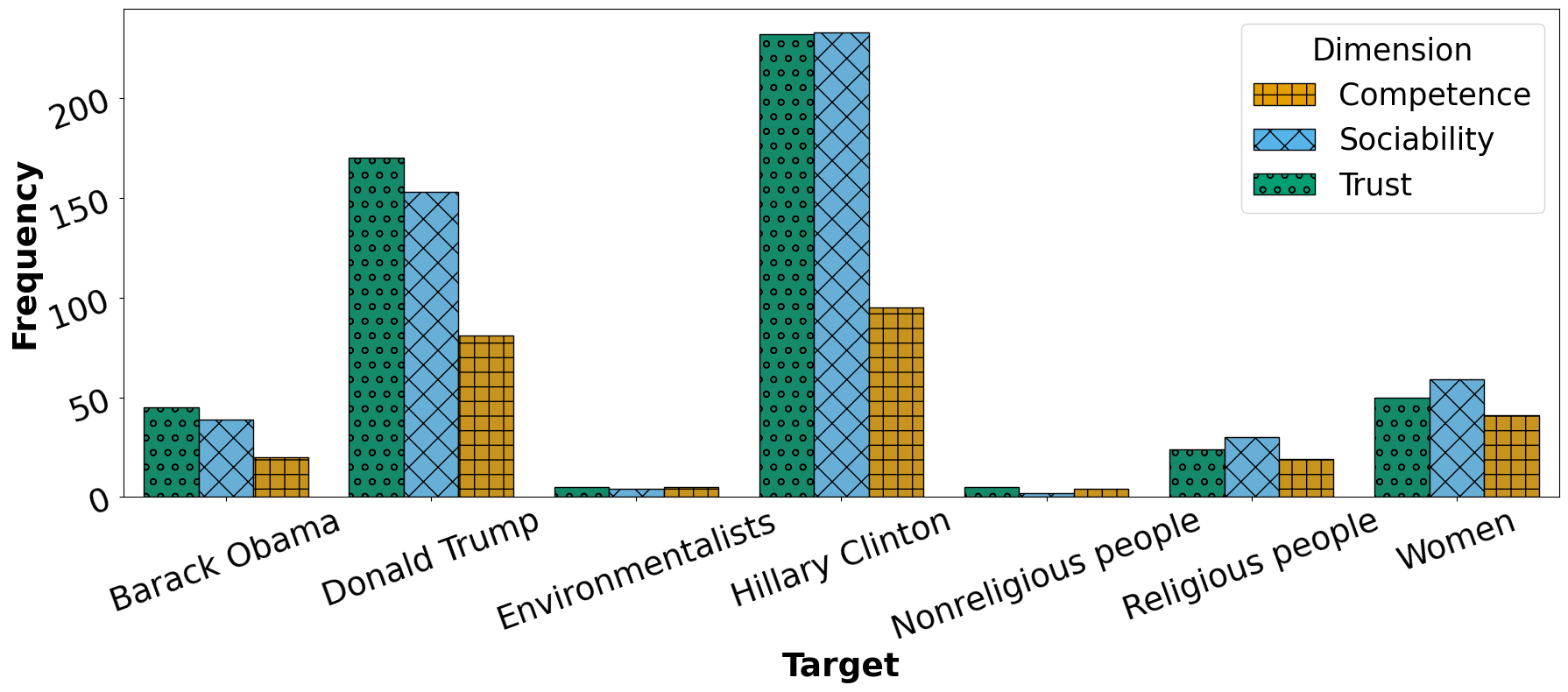}
    \caption{Distribution of \textbf{fine-grained} unanimous judgments across targets and dimensions. The figure shows higher agreement for sentences targeting Clinton and Trump in the trust dimension, and no occurrences for the target Environmentalists.}
\label{fig:unanimous-fine}
\end{figure}

\subsubsection{Soft Unanimity}

This refers to cases where annotators agreed on the overall polarity (low, neutral, or high) of a sentence-target pair even if they did not select the exact same fine-grained score. Annotators reached soft agreement on 1,316 instances (sentence-target pairs), accounting for 26.9\% of all sentence-target pairs in W\&C-Sent. The majority of soft agreement instances were on negative judgments.

\begin{figure}[hbt!]
    \centering
    \includegraphics[width=1\linewidth]{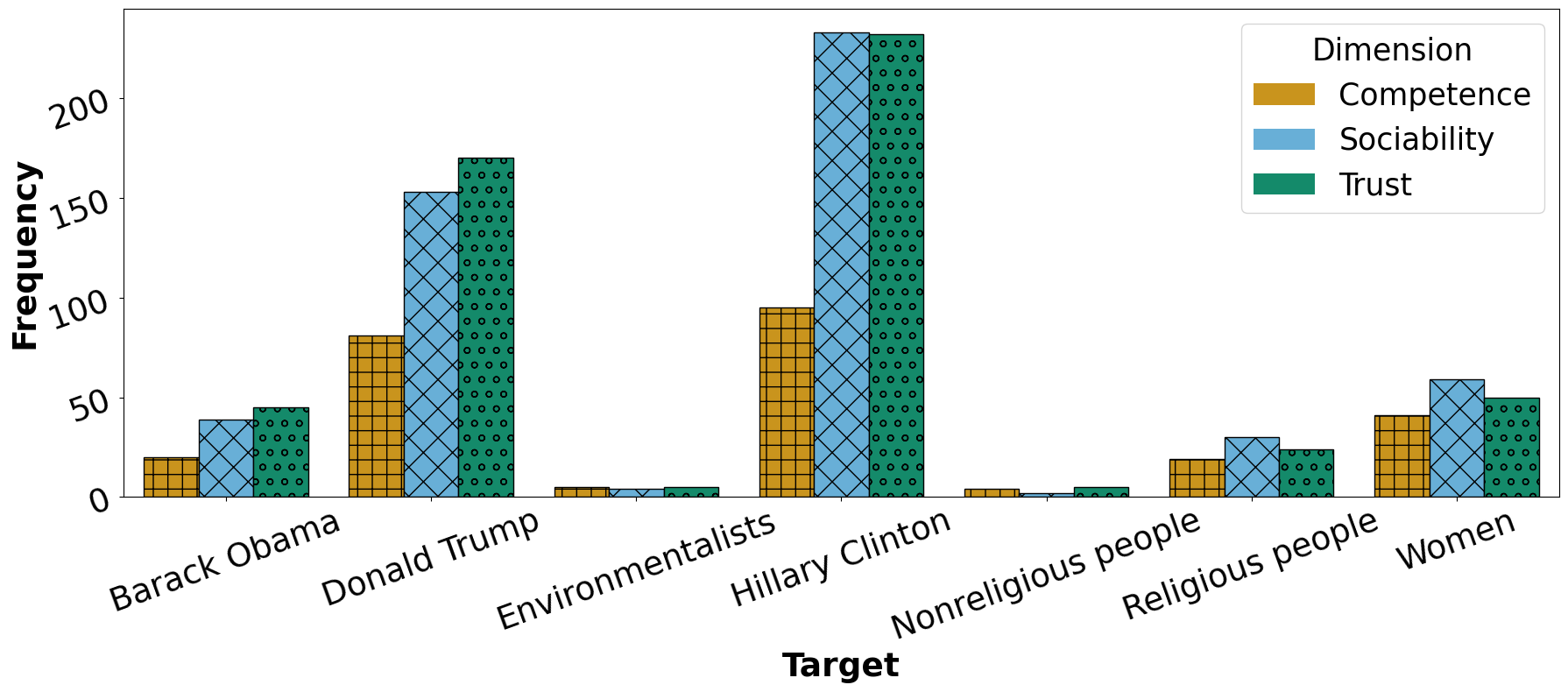}
    \caption{The distribution of \textbf{soft} unanimous judgments across targets and dimensions. The figure higher soft unanimous agreement for the trust and sociability of Clinton and Trump.}
    \label{fig:unanimous-coarse}
\end{figure}

\section{Correlations Between Dimensions}

\label{appendix:corrs}

\begin{figure}[h!]
\centering

\begin{subfigure}{\linewidth}
  \centering
  \includegraphics[width=\linewidth]{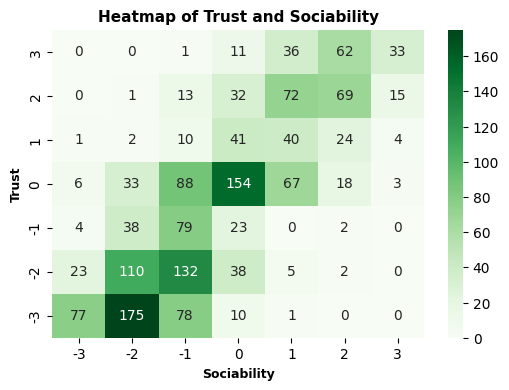}
  \caption{Trust \& Sociability}
\end{subfigure}

\vspace{0.4cm}

\begin{subfigure}{\linewidth}
  \centering
  \includegraphics[width=\linewidth]{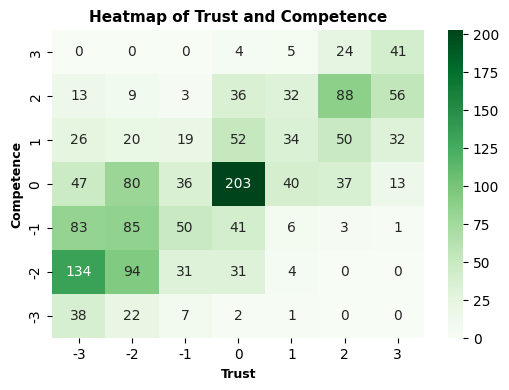}
  \caption{Trust \& Competence}
\end{subfigure}

\vspace{0.4cm}

\begin{subfigure}{\linewidth}
  \centering
  \includegraphics[width=\linewidth]{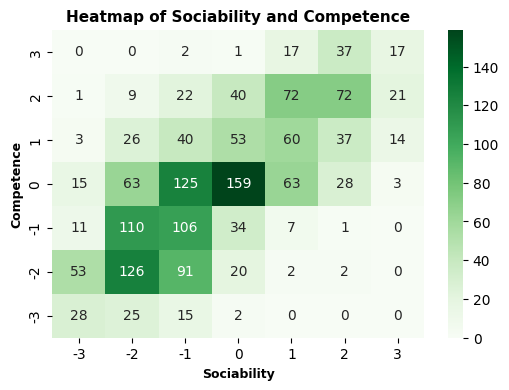}
  \caption{Sociability \& Competence}
\end{subfigure}

\caption{Heat maps of correlation matrices for each pair of dimensions.}
\label{fig:heatmaps}
\end{figure}

Figure \ref{fig:heatmaps} shows three heat maps that show frequent co-occurrence between median neutral scores; meaning, when competence within a sentence is judged as neutral, trust and sociability judgments tend to be neutral in it, too. In addition, they all demonstrate strong co-occurrences around the diagonal, with the highest being between high distrust and moderately high unsociability (-3, -2) in Grid A, followed by high distrust and moderately high incompetence (-3, -2) in Grid B, and moderately high distrust and moderately high unsociability (-2, -2) in Grid C. 

\section{More on Inter-Annotator Agreement}

\subsection{Krippendorff's Alpha}

When broken down by both target and dimension, Krippendorff's $\alpha$ scores show that agreement varies considerably depending on the target of the sentence. The strongest reliability is observed for trust perceptions of Donald Trump ($\alpha = 0.67$). Trust perceptions of Hillary Clinton came second ($\alpha = 0.61$). The scores for sociability perceptions of these two political figures remain weak but higher than for other targets, at around 0.53-0.55. By contrast, Barack Obama's $\alpha$s show weaker consistency, with trust at 0.50 and sociability at 0.48. 

Notably, agreement drops sharply when annotators assessed social groups such as Women, Religious People, Non-religious People, and Environmentalists, where none of the dimensions exceed 0.41 and several fall below 0.30. This suggests that annotators shared more consistent interpretations when evaluating singular, high-profile political figures than collective or socially non-monolithic categories.

Competence scores are consistently poor across all targets. Even for figures like Donald Trump ($\alpha = 0.38$) and Hillary Clinton ($\alpha = 0.32$), agreement on their competence remains weak, while Barack Obama and group-based targets fall even lower. The lowest values are seen for Religious People, Environmentalists, and Nonreligious People. 

\subsection{Split-Half Reliability}

\begin{figure}
    \centering
    \includegraphics[width=1\linewidth]{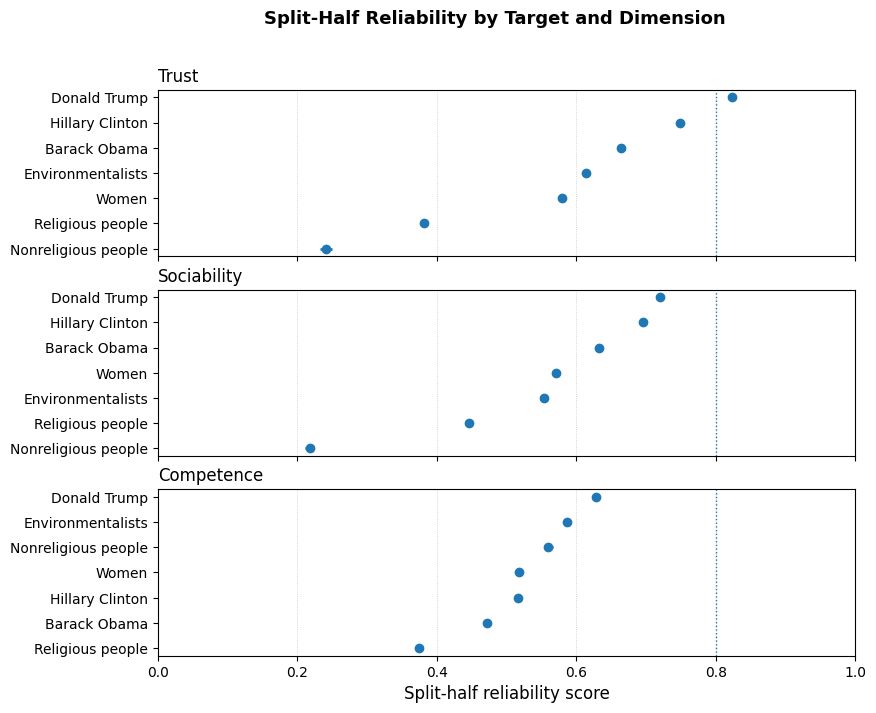}
    \caption{The SHR scores of each target, in each relevant dimension. For each sub-plot, the targets were ordered based on their score, from high to low. The vertical dotted line is the 0.8 reliability cutoff, which only the sentences whose target was Donald Trump cross in the trust dimension.}
    \label{fig:shr-target-dim}
\end{figure}

The process by which SHR scores were computed for each dimension was repeated again for each target. Figure \ref{fig:shr-target-dim} illustrates these scores, showing some interesting highlights. The sentences whose target was Donald Trump showed the highest stability across all three dimensions, and especially in the trust dimension where the SHR score exceeds 0.8. The targets Hillary Clinton and Barack Obama fall right behind Donald Trump in trust and sociability but had very poor SHR in competence. The target Women had mediocre stability across all three dimensions, while Religious People, Non-religious People, and Environmentalists came last in almost all dimensions, probably due to their small size in the dataset.

\label{appendix:IAA}

\section{Regression Models}

\subsection{Logistic Regression using TF-IDF Features}

A logistic regression model, using TF-IDF features and the mean of annotator scores as the output, was trained first. 
Table \ref{tab:regression-table-with-lr} shows that the TF-IDF models perform markedly worse, which means that simple lexical representations are inadequate for detecting T/S/C in text. This is especially seen in the much lower correlations between its predictions and the real scores, proving the importance of the contextual BERT representations.

\begin{table}[hbt!]
\centering
\small
\resizebox{\columnwidth}{!}{%
\begin{tabular}{|c|c|cccccc|}
\hline
\textbf{Dimension} & \textbf{Metric}        & \textbf{MAE} & \textbf{RMSE} & \textbf{$\rho$} & \textbf{Acc.} & \textbf{F1} & \textbf{$\pm1$ Acc.} \\ 
\hline
\multirow{3}{*}{Trust} & \underline{TF-IDF} & 1.36          & 1.90          & 0.44             & 29.7          & 29.5      & 65.1 \\ 
% \cline{2-8}
                       & \underline{Median} & 1.01           & 1.40          & 0.76             & 35.8          & \textbf{31.2}      & 76.5 \\ 
% \cline{2-8}
                      & \underline{Mean}   & \textbf{0.81}         & \textbf{1.10}     & \textbf{0.77}    & \textbf{37.0}   & 29.4   & \textbf{82.6} \\  
\hline
% Rows 5-7: second multirow block
\multirow{3}{*}{Sociability} & \underline{TF-IDF} & 1.18           & 1.70         & 0.43             & 31.2          & 26.2      & 70.3 \\ 
% \cline{2-8}
                            & \underline{Median} & 0.83           & 1.03          & 0.76             & 35.5          & 25.7      & 85.0 \\ 
% \cline{2-8}
                            & \underline{Mean}   & \textbf{0.66}  & \textbf{0.86}    & \textbf{0.78}   & \textbf{45.6}    & \textbf{33.3}   & \textbf{88.7} \\ 
\hline
% Rows 8-10: third multirow block
\multirow{3}{*}{Competence} & \underline{TF-IDF} & 1.40           & 1.80         & 0.35             & 19.0          & 15.8      & 61.5 \\ 
% \cline{2-8}
                            & \underline{Median} & 1.02           & 1.29          & 0.60     & 28.7          & 21.7      & 77.7 \\ 
% \cline{2-8}
                            & \underline{Mean}   & \textbf{0.83}           & \textbf{1.06}    & \textbf{0.63}     & \textbf{37.3}   & \textbf{25.2}   & \textbf{85.3} \\  
\hline
\end{tabular}}
\caption{The results of the seven-class regression models for each dimension. The results for the best variant are in \textbf{bold}. This table highlights how using TF-IDF embeddings do not perform well compared to the contextual BERT-based embeddings.}
\label{tab:regression-table-with-lr}
\end{table}

\subsection{Seven-Class Regression}

The Huber loss was chosen for a loss function due the advantages mentioned earlier: it balances the sensitivity of MSE to outliers with the stability of MAE for an optimized performance. In addition, input examples that would be used to train the models were structured as paired sequences, where the ``Text" and ``Target" columns were jointly tokenized using the BERTweet tokenizer. Later, W\&C-Sent was partitioned into training, validation, and test sets using grouped splits based on text instances; identical texts, of which there is a lot in the dataset but with different targets, were specifically designed to not appear across different data splits, thereby helping avoid bias and inflated performance. 

Training was conducted using the ``Trainer" class from Hugging Face \citep{huggingface}. Model performance was assessed with MAE, RMSE, and Spearman correlation. Two variants of these experiments were run, one using the median score and one using the mean as the target label. This was motivated by pure curiosity, as I was wondering whether regression models would perform better using the mean or the median. After the training was completed, each model was assessed on the same held-out test set, and its continuous predictions were rounded into seven discrete ordinal categories to ease comparison between the variants.

The model for each dimension was evaluated using:

\begin{enumerate}
    \item MAE and RMSE, for error magnitude
    \item Spearman $\rho$, for correlation with human ratings
    \item Accuracy, to observe the exact matches
    \item Macro F1 score, which measures the overall balance between precision and recall across all classes
    \item And the within-1-bin accuracy, which measures how often model predictions fall within one rating level of the true labels captures near-miss performance in ordinal tasks
\end{enumerate}

\label{appendix:seven-class}

\section{Classification Models}

\subsection{Experimental Setup}

Due to class imbalance, a custom subclass of the Trainer class was implemented and in which a weighted negative log-likelihood loss was defined; this involved class weights to be derived from the inverse of the class frequencies in the training data, to ensure that the underrepresented neutral classes exerted a proportionally stronger influence on the weight optimization process. This also ensures better generalization across all three categories.

Input processing mirrored that of the regression setup. Paired sequences of ``Text" and ``Target" were tokenized jointly with the BERTweet tokenizer and truncated to the same fixed maximum length. Additionally, W\&C-Sent was partitioned into training, validation, and test sets the same way, too. 

The training was carried out identically as well, but the evaluation metrics changed due to the type of the model. Evaluation here involved accuracy and macro-averaged F1 score, with the F1 score being the metric by which the best model was decided; F1 emphasizes balanced performance across classes, particularly due to the imbalanced label distributions.

Two versions of this experiment was run, one included coarsening the W\&C-Sent dataset using majority-vote coarsening while the second followed the method used in \cite{wordsofwarmth}. The class distribution when the labels are coarsened using first method is shown in Table \ref{tab:coarse-labels}.

\label{appendix:three-class}

\section{Hyper-parameters for Code Reproducibility}

\subsection{Classification}

\begin{table}[h!]
\centering
\resizebox{\columnwidth}{!}{%
\begin{tabular}{ll}
\toprule
\textbf{Component} & \textbf{Value} \\
\midrule
Task & Three-class classification (text + target pair) \\
Base model & \texttt{vinai/bertweet-base} \\
Tokenization & Truncate + pad to fixed length \\
Max sequence length & 256 \\
Inputs & Text as primary, Target as text\_pair \\
Labels & Median score, mapped to a numeric value \\
Num classes & 3 \\
Class weighting & Inverse frequency on training split, used in CrossEntropyLoss \\
Split method & GroupShuffleSplit (GSS) on Text\\
Test size & 0.2 \\
Validation size & 0.2 (using a second GSS) \\
Random seed & 13 \\
Epochs & 10 \\
Train batch size & 16 \\
Eval. batch size & 32 \\
Learning rate & $2\times10^{-5}$ \\
Evaluation cadence & Each epoch \\
Model selection & Using macro-F1 \\
Early stopping & Patience with 3 epochs \\
Reported metrics & Accuracy, macro-F1 (on test set) \\
Per-dimension setting & Separate model per dimension \\
\bottomrule
\end{tabular}}
\caption{Hyperparameters and setup for the 3-class classifier.}
\label{tab:clf_hparams}
\end{table}

\begin{table}[h!]
\centering
\resizebox{\columnwidth}{!}{%
\begin{tabular}{ll}
\toprule
\textbf{Component} & \textbf{Value} \\
\midrule
Task & Regression on text + target pair \\
Base model & \texttt{vinai/bertweet-base} \\
Tokenization & Truncate + pad to fixed length \\
Max sequence length & 256 \\
Inputs & Text as primary, Target as text\_pair \\
Target column & Median or mean score \\
Loss & Huber (SmoothL1) with $\beta=1.0$ \\
Split method & GroupShuffleSplit (GSS) on the Text column \\
Test size & 0.2 \\
Validation size & 0.2 (using a second GSS) \\
Random seed & 13 \\
Epochs & 10 \\
Train batch size & 16 \\
Eval batch size & 32 \\
Learning rate & $1\times10^{-5}$ \\
Weight decay & 0.01 \\
Warmup ratio & 0.06 \\
Model selection & Best validation MAE \\
Early stopping & Patience with 3 epochs \\
Logging & Every 50 steps \\
Reported test metrics & MAE, RMSE, Spearman $\rho$ \\
Per-dimension setting & Separate model per dimension \\
\bottomrule
\end{tabular}
}
\caption{Hyperparameters and setup for the regression model.}
\label{tab:reg_hparams}
\end{table}

\newpage 
\textcolor{white}{-}
\newpage

\section{Full Results with Precision and Recall Scores}

\label{appendix:prec-recall-scores}
\begin{table}[hbt!]
\centering
\small
\resizebox{\columnwidth}{!}{%
\begin{tabular}{llcccccc}
\toprule
\textbf{Dimension} & \textbf{Model} & \textbf{Accuracy} & \textbf{F1} &\textbf{ $\pm1$ Accuracy} & \textbf{Precision} & \textbf{Recall} \\
\midrule
\multirow{15}{*}{Trust}
& Dummy Classifier               & 0.22 & 0.08 & 0.58 & 0.05  & 0.22 \\
& TF-IDF Regression              & 0.30 & 0.30 & 0.65 & 0.31  & 0.30 \\
& Seven-Class Regression          & 0.35 & 0.31 & 0.83 & 0.43  & 0.32 \\
& Gemma3 ZS                       & 0.36 & 0.34 & 0.78 & 0.36  & 0.36 \\
& Gemma3 FS                       & 0.38 	 & 	 0.33 	 & 	 0.79 	 & 	 0.37 	 & 	 0.35 \\
& Qwen2.5 ZS                        & 0.27 & 0.25 & 0.72 & 0.29  & 0.29 \\
& Qwen2.5 FS                        & 0.35 	 & 	 0.35 	 & 	 0.76 	 & 	 0.38 	 & 	 0.4 \\
& GPT-4o ZS                      & 0.43 & 0.42 & 0.91 & 0.43  & 0.50 \\
& GPT-4o FS                      & 0.39 	 & 	 0.4 	 & 	 0.84 	 & 	 0.4 	 & 	 0.43 \\
& GPT-4o-mini ZS                 & 0.27 & 0.26 & 0.60 & 0.32  & 0.36 \\
& GPT-4o-mini FS                 & 0.2 	 & 	 0.17 	 & 	 0.54 	 & 	 0.25 	 & 	 0.19 \\
& GPT-5.2 ZS                  & 0.41 & 0.38 & 0.87 & 0.45 & 0.41 \\
& GPT-5.2 FS      & 0.40 & 0.39 & 0.89 & 0.45 & 0.40 \\
& Qwen3 ZS                    & 0.32 & 0.30 & 0.78 & 0.33 & 0.31 \\
& Qwen3 FS     & 0.30 & 0.26 & 0.78 & 0.32 & 0.28 \\
\midrule
\multirow{15}{*}{Sociability}
& Dummy Classifier               & 0.26  & 0.11 & 0.67 & 0.07 & 0.26 \\
& TF-IDF                         & 0.31 & 0.26 & 0.70 & 0.26 & 0.27 \\
& Seven-Class Regression          & 0.46  & 0.34 & 0.88 & 0.33 & 0.35 \\
& Gemma3 ZS                       & 0.34  & 0.27 & 0.82 & 0.34 & 0.33 \\
& Gemma3 FS                       & 0.31 	 & 	 0.23 	 & 	 0.76 	 & 	 0.29 	 & 	 0.27 \\
& Qwen2.5 ZS                        & 0.20  & 0.20 & 0.69 & 0.31 & 0.28 \\
& Qwen2.5 FS                        &0.25 	 & 	 0.25 	 & 	 0.67 	 & 	 0.29 	 & 	 0.28 \\
& GPT-4o ZS                      & 0.44  & 0.40 & 0.92 & 0.42 & 0.43 \\
& GPT-4o FS                      & 0.38 	 & 	 0.35 	 & 	 0.87 	 & 	 0.36 	 & 	 0.42 \\
& GPT-4o-mini ZS                 & 0.13  & 0.14 & 0.52 & 0.31 & 0.31 \\
& GPT-4o-mini FS                 & 0.24 	 & 	 0.2 	 & 	 0.59 	 & 	 0.27 	 & 	 0.24 \\
& GPT-5.2 ZS             & 0.31 & 0.30 & 0.83 & 0.45 & 0.32 \\
& GPT-5.2 FS  & 0.40 & 0.37 & 0.88 & 0.44 & 0.36 \\
& Qwen3 ZS               & 0.35 & 0.30 & 0.81 & 0.39 & 0.31 \\
& Qwen3 FS   & 0.31 & 0.24 & 0.82 & 0.33 & 0.32 \\
\midrule
\multirow{15}{*}{Competence}
& Dummy Classifier               & 0.24 & 0.09 & 0.60 & 0.06 & 0.24 \\
& TF-IDF                         & 0.19 & 0.16 & 0.62 & 0.18 & 0.18 \\
& Seven-Class Regression          & 0.36 & 0.24 & 0.86 & 0.26 & 0.26 \\
& Gemma3 ZS                       & 0.22 & 0.19 & 0.58 & 0.32 & 0.24 \\
& Gemma3 FS                       & 0.35 	 & 	 0.27 	 & 	 0.7 	 & 	 0.32 	 & 	 0.33 \\
& Qwen2.5 ZS                        & 0.22 & 0.19 & 0.59 & 0.34 & 0.17 \\
& Qwen2.5 FS                        & 0.30 	 & 	 0.28 	 & 	 0.65 	 & 	 0.31 	 & 	 0.28 \\
& GPT-4o ZS                      & 0.34 & 0.31 & 0.80 & 0.34 & 0.30 \\
& GPT-4o FS                      & 0.24 	 & 	 0.24 	 & 	 0.64 	 & 	 0.24 	 & 	 0.3 \\
& GPT-4o-mini ZS                 & 0.09 & 0.11 & 0.37 & 0.28 & 0.18 \\
& GPT-4o-mini FS                 & 0.22 	 & 	 0.2 	 & 	 0.6 	 & 	 0.27 	 & 	 0.24 \\
& GPT-5.2 ZS            & 0.28 & 0.25 & 0.73 & 0.33 & 0.26 \\
& GPT-5.2 FS & 0.27 & 0.24 & 0.71 & 0.30 & 0.24 \\
& Qwen3 ZS              & 0.22 & 0.18 & 0.60 & 0.31 & 0.19 \\
& Qwen3 FS   & 0.25 & 0.21 & 0.67 & 0.34 & 0.22 \\

\bottomrule

\end{tabular}
}
\caption{Performance of regression models and LLMs on fine-grained prediction.}
\label{tab:fine-prec-recall}
\end{table}

% ------------------------

\begin{table}[hbt!]
\centering
\small
\resizebox{\columnwidth}{!}{%
\begin{tabular}{llccccc}
\toprule
\textbf{Dimension} & \textbf{Model} & \textbf{Accuracy} & \textbf{F1} & \textbf{Precision} & \textbf{Recall} \\
\midrule
\multirow{9}{*}{Trust}
& BERT Classification    & 0.79 & 0.59 & 0.598 & 0.579 \\
% & Mean-based Classification       & 0.75 & 0.61  0.63  & 0.62 \\
& Gemma ZS                & 0.74 & 0.60 & 0.58  & 0.64 \\
& Gemma FS                & 0.72 & 0.60 & 0.58  & 0.67 \\
& Qwen2.5 ZS                 & 0.62 & 0.53 & 0.60  & 0.70 \\
& Qwen2.5 FS                 & 0.61 & 0.51 & 0.57  & 0.69 \\
& GPT-4o-ZS               & 0.86 & 0.78 & 0.79  & 0.78 \\
& GPT-4o-FS               & 0.77 & 0.71 & 0.75  & 0.73 \\
& GPT-4o-mini ZS          & 0.78 & 0.65 & 0.63  & 0.70 \\
& GPT-4o-mini FS          & 0.70 & 0.58 & 0.57  & 0.61 \\
& GPT-5.2 ZS &  0.81 	 & 	 0.72 	 & 	 0.98 	 & 	 0.73 	 & 	 0.71  \\
& GPT-5.2 FS &  0.78 	 & 	 0.71 	 & 	 0.99 	 & 	 0.73 	 & 	 0.71  \\
& Qwen3 ZS &  0.79 	 & 	 0.66 	 & 	 0.94 	 & 	 0.65 	 & 	 0.69   \\
& Qwen3 FS &   0.72 	 & 	 0.62 	 & 	 0.94 	 & 	 0.62 	 & 	 0.62  \\

\midrule

\multirow{9}{*}{Sociability}
& BERT Classifier    & 0.81  & 0.54 & 0.52 & 0.57 \\
% & Mean-based Classification       & 0.72  & 0.66 & 0.68 & 0.67 \\
& Gemma ZS                & 0.69  & 0.56 & 0.55 & 0.58 \\
& Gemma FS                & 0.71  & 0.54 & 0.53 & 0.61 \\
& Qwen2.5 ZS                 & 0.47  & 0.42 & 0.44 & 0.59 \\
& Qwen2.5 FS                 & 0.34  & 0.34 & 0.44 & 0.65 \\
& GPT-4o-ZS               & 0.80  & 0.71 & 0.72 & 0.71 \\
& GPT-4o-FS               & 0.72  & 0.66 & 0.69 & 0.71 \\
& GPT-4o-mini ZS          & 0.74  & 0.54 & 0.56 & 0.57 \\
& GPT-4o-mini FS          & 0.62  & 0.47 & 0.48 & 0.48 \\
& GPT-5.2 ZS &  0.83 	 & 	 0.74 	 & 	 0.98 	 & 	 0.75 	 & 	 0.73  \\
& GPT-5.2 FS & 0.79 	 & 	 0.72 	 & 	 0.99 	 & 	 0.74 	 & 	 0.73   \\
& Qwen3 ZS &  0.74 	 & 	 0.59 	 & 	 0.91 	 & 	 0.58 	 & 	 0.61   \\
& Qwen3 FS &  0.67 	 & 	 0.58 	 & 	 0.96 	 & 	 0.58 	 & 	 0.6   \\

\midrule

\multirow{9}{*}{Competence}
& BERT Classifier     & 0.74 & 0.50 & 0.49 & 0.51 \\
% & Mean-based Classification       & 0.58 & 0.55 & 0.56 & 0.55 \\
& Gemma ZS                & 0.52 & 0.46 & 0.47 & 0.47 \\
& Gemma FS                & 0.52 & 0.43 & 0.46 & 0.44 \\
& Qwen2.5 ZS                 & 0.47 & 0.47 & 0.49 & 0.61 \\
& Qwen2.5 FS                 & 0.56 & 0.53 & 0.53 & 0.59 \\
& GPT-4o-ZS               & 0.63 & 0.59 & 0.59 & 0.60 \\
& GPT-4o-FS               & 0.60 & 0.57 & 0.57 & 0.60 \\
& GPT-4o-mini ZS          & 0.53 & 0.40 & 0.46 & 0.44 \\
& GPT-4o-mini FS          & 0.47 & 0.37 & 0.42 & 0.52 \\
& GPT-5.2 ZS &  0.49 	 & 	 0.46 	 & 	 0.86 	 & 	 0.51 	 & 	 0.51  \\
& GPT-5.2 FS & 0.46 	 & 	 0.43 	 & 	 0.82 	 & 	 0.5 	 & 	 0.49   \\
& Qwen3 ZS &  0.64 	 & 	 0.56 	 & 	 0.84 	 & 	 0.57 	 & 	 0.55   \\
& Qwen3 FS &  0.67 	 & 	 0.58 	 & 	 0.9 	 & 	 0.59 	 & 	 0.58   \\

\bottomrule
\end{tabular}}
\caption{Performance of BERT classification model and LLMs on coarse-grained prediction.}
\label{tab:coarse-prec-recall}
\end{table}

\newpage
\section{LLM Prompts}

\label{appendix:llm-prompts}

\subsection{System Message}

``You are a professional language analyst and sociologist.\newline
You work on warmth and competence, which are two concepts in sociology.\newline
You know that warmth is decomposed into trust, which relates to the moral and personal aspect of the target; and sociability, which relates to the relational and societal aspect and impact of the target.\newline
You work on a sentence-level: you read the full sentence, with all its components, before you assess it.\newline
There is a degree of subjectivity in this task, so you consider the meaning that the general population would agree with and consider.\newline
You do not add information that does not appear in the text.\newline
The context matters the most to you. You consider all relevant information.\newline
You understand that sarcasm is an ironic remark meant to mock by saying something different than what the speaker really means.\newline
You understand that irony is the humorous or mildly sarcastic use of words to imply the opposite of what they normally mean.\newline
You understand that hyperbole is the extreme, dramatic exaggeration for effect, not meant to be taken literally.\newline
You understand how irony, sarcasm, and hyperbole are employed in social media and in slang, and that the literal wording does not always reflect the true meaning and sentiment.\newline
You assess warmth and competence on a scale from -3 to +3\footnote{Coarse-grained prompts replaced this line with ``You assess warmth and competence on the scale of ``low", ``neutral, not applicable, not expressed", and ``high".
"}.\newline
You adhere to the provided target. Meaning, you understand that even if the speaker is explicitly expressing opinions towards entity\_X, you give the score for trust towards entity\_Y only if the target given to you is entity\_Y.\newline
You do not use first-person pronouns such as ``I" or ``we" in your answer.\newline
You adhere to the targets given to you in your assessment of warmth or competence.\newline
You understand that the use of hashtags in social media can be complementary to the text. However, that is not always the case, as hashtags can be used to categorize content, increase visibility, and connect users with relevant discussions and communities, rather than as sentiment markers.\newline
You understand that when the target is ``religious people", then it refers to religions and those who believe in any God or practice any religion.\newline
You understand that when the target is ``nonreligious people", then it refers to atheism and those who are not religious (i.e., atheists/agnostics).\newline
You understand that when the target is ``women", then it refers to women or girls and/or interconnected topics: sexism, feminism, misogyny, bias, and female representation.\newline
You understand that when the target is ``Hillary Clinton", then it refers to the 2016 US presidential candidate and former Secretary of State Hillary Clinton.\newline
You understand that when the target is ``Donald Trump", then it refers to the 2016 US presidential candidate and current US president Donald Trump.\newline
You understand that when the target is ``Barack Obama", then it refers to the former US president Barack Obama.\newline
You understand that when the target is ``Environmentalists", then it refers to to environment and climate change activists."\newline

\label{appendix:system-message}

\subsection{Fine-Grained, Zero-Shot User Role Messages}

\subsubsection{Trust}

``Some context:\newline
Social psychology research has shown that individuals rapidly and subconsciously evaluate others, groups, and even themselves along the dimensions of warmth (W) and competence (C).\newline
Recently, psychologists have modelled warmth through two dimensions: trust (T) and sociability (S).\newline
Your task is to assess the degree of trust towards a specific target within a sentence.\newline
The focus in this dimension is on the personal / moral aspect of the target.\newline
High trust can be defined as morality, kindness, sincerity, trustworthiness, and honesty.\newline
Words associated with high trust: charity, mother, compliment, affectionate.\newline
Low trust can be defined as immorality, insincerity, dishonesty, untrustworthiness, dubiousness, and maliciousness.\newline
Words associated with low trust: discredit, bribe, espionage, disinformation, disloyal.\newline

\noindent \underline{Instructions}:\newline
Consider the entire meaning of the sentence before attempting to give the relevant scores.\newline
You will be given a text snippet and a target (group, entity, or individual).\newline
Rate the apparent level of trust that the sentence's author seems to express towards the specified target.\newline
Choose one label from these seven labels:\newline
high distrust\newline
moderate distrust\newline
slight distrust\newline
neutral, not applicable, not expressed\newline
slight trust\newline
moderate trust\newline
high trust\newline

\noindent Here is the sentence: \{  \}.\newline
Its target is: \{   \}.\newline

\noindent In order to assess trust, try to answer the following questions:\newline
1. What is the degree of trust towards \{   \} that the author of the text seems to express?\newline
2. Does the author seem to perceive \{   \} as trustworthy or untrustworthy / moral or immoral / honest or dishonest?\newline
Remember: even if the speaker is explicitly targeting someone else, since the target is \{   \}, your score should be an assessment of the trust towards \{   \} only.\newline
In the format of a JSON file or a Python dictionary, you should provide your justification saved in a key called ``reason". Then, based on your justification, add your rating to a key called ``label"."\newline

\label{appendix:trust-message}

\subsubsection{Sociability}

Some context:\newline
Social psychology research has shown that individuals rapidly and subconsciously evaluate others, groups, and even themselves along the dimensions of warmth (W) and competence (C).\newline
Recently, psychologists have modelled warmth through two dimensions: trust (T) and sociability (S).\newline
Your task is to assess the degree of sociability towards a specific target within a sentence.\newline
The focus in this dimension is on the social aspect of the target and its relational impact on others or society as a whole.\newline
High sociability can be defined as friendliness, sociableness, generosity, and helpfulness.\newline
Words associated with high sociability: helpful, intimate, laugh, celebration, reliant, entertain, social club, bestie.\newline
Low sociability can be defined as antisocial behavior, lack of generosity, inconsiderateness, indifference, and unhelpfulness.\newline
Words associated with low sociability: ingrate, abduct, selfish, theft, egomaniacal, pervert.\newline
\underline{Instructions}:\newline
Consider the entire meaning of the sentence before attempting to give the relevant scores.\newline
You will be given a text snippet and a target (group, entity, or individual).\newline
Rate the apparent level of sociability that the sentence's author seems to express towards the specified target.\newline
Choose one label from these seven labels:\newline
high unsociability\newline
moderate unsociability\newline
slight unsociability\newline
neutral, not applicable, not expressed, etc.\newline
slight sociability\newline
moderate sociability\newline
high sociability\newline

\noindent Here is the sentence: \{    \}.\newline
Its target is: \{    \}.

\noindent In order to assess sociability, try to answer the following questions:\newline
1. What is the degree of sociability towards \{    \} that the author of the text seems to express? \newline
2. Does the author seem to perceive \{    \} as sociable or antisocial? Helpful or unhelpful?\newline
Remember: even if the speaker is explicitly targeting someone else, since the target is \{    \}, your score should be an assessment of perceived sociability trust towards \{    \} only.\newline
In the format of a JSON file or a Python dictionary, you should provide your justification saved in a key called ``reason". Then, based on your justification, add your rating to a key called ``label"."

\label{appendix:soc-message}

\subsubsection{Competence}

Some context:\newline
Social psychology research has shown that individuals rapidly and subconsciously evaluate others, groups, and even themselves along the dimensions of warmth (W) and competence (C).\newline
Your task is to assess the degree of competence towards a specific target within a sentence.\newline
Competence, in the broader sense, is interchangeable with `ability' or `capability'.
Competence can be defined as ability, power, dominance, being in control, importance, having influence, and assertiveness.\newline
Words associated with high competence: hitman, heroical, entrepreneurship, strategies, superman, viper, impunity.\newline
Incompetence can be defined as submissiveness, not being in control of a situation, being controlled or guided by outside factors, or weakness.\newline
Words associated with low competence: bootlicker, talentless, crash landing, bedridden, underfed.\newline
\underline{Instructions}:\newline
Consider the entire meaning of the sentence before attempting to give the relevant scores.\newline
You will be given a text snippet and a target (group, entity, or individual).\newline
Rate the apparent level of competence that the sentence's author seems to express towards the specified target.\newline
Choose one label from these seven labels:
high incompetence\newline
moderate incompetence\newline
slight incompetence\newline
neutral, not applicable, not expressed\newline
slight competence\newline
moderate competence\newline
high competence\newline
Adhere to the literal meaning of competence, which may be ``positive" (e.g., a CEO) or ``negative" (e.g., a villain or a dictator). Both types are considered competence, regardless of the outcomes.\newline

\noindent Here is the sentence: \{    \}.\newline
Its target is: \{    \}.

\noindent In order to assess competence, try to answer the following questions:\newline
1. What is the degree of competence towards \{    \} that the author of the text seems to express? \newline
2. Does the author seem to perceive \{    \} in control or out of control? Active or passive? Powerful or weak?\newline
Remember: even if the speaker is explicitly targeting someone else, since the target is \{    \}, your score should be an assessment of the competence towards \{    \} only.\newline
In the format of a JSON file or a Python dictionary, you should provide your justification saved in a key called ``reason". Then, based on your justification, add your rating to a key called ``label".

\label{appendix:comp-message}

\subsection{Fine-Grained, Few-Shot User Role Messages}

\subsubsection{Trust}

Some context:\newline
Social psychology research has shown that individuals rapidly and subconsciously evaluate others, groups, and even themselves along the dimensions of warmth (W) and competence (C).\newline
Recently, psychologists have modelled warmth through two dimensions: trust (T) and sociability (S).\newline
Your task is to assess the degree of trust towards a specific target within a sentence.\newline

\noindent The focus in this dimension is on the personal / moral aspect of the target.\newline
High trust can be defined as morality, kindness, sincerity, trustworthiness, and honesty.\newline
Words associated with high trust: charity, mother, compliment, affectionate.
Low trust can be defined as immorality, insincerity, dishonesty, untrustworthiness, dubiousness, and maliciousness.\newline
Words associated with low trust: discredit, bribe, espionage, disinformation, disloyal.

\noindent Here are examples, each containing the text, the target, the score, and the rationale behind the score.\newline

\noindent Example 1:\newline

\noindent \underline{Target}: Women\newline
\underline{Text}: My wife is the most caring person I've ever met ... she's the only woman in a house full of testosterone . She never stops working whether it's at home or being an RN . I cant keep up but I try . She makes me a better person . I'd be lost without her . Oh and she's smoking hot too.\newline
\underline{Trust score}: +3 (high trust). \newline
\underline{Rationale}: The author expresses maximum trust in women through his wife as a representative example. He portrays his wife as a trustworthy, dependable, and caring individual who is essential to his well-being ("I'd be lost without her"). The statement ``she makes me a better person" implies the author views his wife as having a strong moral character that positively influences him.\newline

\noindent Example 2:\newline

\noindent \underline{Target}: Women\newline
\underline{Text}: ``when i was 16 i had a folder of “ feel good songs ” and everyday i would select one and send it to my best friend along with a paragraph of how much they meant to me and why i they should be happy and then i would lay in bed thinking “ no i’m not gay this is just what girls do :) ”"\newline
\underline{Trust score}: +3 (high trust).\newline
\underline{Rationale}: The author attributes sincere, genuine intentions to women through the described behaviors. The daily practice of sending heartfelt messages ``about how much they meant to me and why they should be happy" suggests honesty and sincerity in women’s relationships. The author portrays women as having good moral intentions in their friendships by genuinely caring about others' wellbeing. \newline

\noindent Example 3:\newline

\noindent \underline{Target}: Public Office Candidate Maya Thompson\newline
\underline{Text}: I'm really excited for 2026 after finding out the amount @MayaThompson raised. We got this.\newline
\underline{Trust score}: +2 (moderate trust).\newline
\underline{Rationale}: The author expresses confidence and optimism toward the target upon learning about the amount of money she raised, framing this as a positive and reassuring signal. The phrase “We got this” implies alignment with and belief in the target’s reliability and capacity to lead or represent shared goals. The enthusiastic endorsement reflects a meaningful level of confidence in the target’s legitimacy and dependability within the political context.\newline

\noindent Example 4:\newline

\noindent \underline{Target}: Public Office Candidate Alex Rivera \newline
\underline{Text}: I support Jordan Lee because the country needs a new direction, but if Lee loses the primary, I’ll support Alex Rivera.\#Election2024 \newline
\underline{Trust score}: +1 (slight trust).\newline
The author expresses conditional support for Alex Rivera if Jordan Lee loses the primary. This implies a baseline level of acceptance and confidence in Rivera's moral and political legitimacy, even if she is not the author's first choice. The willingness to support her suggests the author does not view her as dishonest, corrupt, or morally unfit.\newline

\noindent Example 5:\newline

\noindent \underline{Target}: Women \newline
\underline{Text}: I need feminism because the United States is one of the only countries that doesn't give paid maternity leave.\newline
\underline{Trust score}: 0 (neutral). \newline
\underline{Rationale}: The author’s statement is focused on policy rather than character traits and doesn't make any attributions about women's trustworthiness or morality. This is a policy statement rather than a personal attribution about women's trustworthiness. This should help you distinguish between advocacy/policy statements and personal/characteristic attributions.\newline

\noindent Example 6:\newline

\noindent \underline{Target}: Religious people \newline
\underline{Text}: Could all those who believe in a god please leave. The meeting will now continue for the grown ups only.\newline
\underline{Trust score}: 0 (neutral).\newline
\underline{Rationale}: The criticism is entirely focused on intellectual maturity rather than character or morality. The author is suggesting that religious people shouldn't participate in this particular discussion. On the other hand, the author doesn't make any claims about religious people's morality, honesty, sincerity, or trustworthiness. There are no accusations of deception, dishonesty, or moral failings. \newline

\noindent Example 7:\newline

\noindent \underline{Target}: Religious people \newline
\underline{Text}: When people use religion as a reason to exclude others, they should not be surprised when others push back against them.\newline
\underline{Trust score}: -1 (slight distrust).\newline
\underline{Rationale}: The statement frames religious people as potential discriminators, suggesting that members of this group may engage in unfair or exclusionary behavior toward others. This implicitly casts doubt on their moral reliability and fairness, which are core components of trust. However, the author criticizes discriminatory behavior associated with the social group but does not assert corruption, deception, or bad faith, which earned it a mild distrust score.\newline

\noindent Example 8:\newline

\noindent \underline{Target}: Public Candidate Riley Chen\newline
\underline{Text}: It sounds like the candidate promised opposite things to different people just to please them.\newline
\underline{Trust score}: -2 (moderate distrust).\newline
\underline{Rationale}: The author directly accuses the target of making inconsistent and contradictory promises to different people about the same policy issue. This frames the target as insincere and politically opportunistic, suggesting a lack of honesty and moral reliability. It implies strategic deception and two-faced behavior, which undermines perceptions of trustworthiness.\newline

\noindent Example 9:\newline
\underline{Target}: Public Office Candidate Jordan Reed\newline
\underline{Text}: Would you want to be in a long-term relationship with someone who hides information and lies to you? Then don’t vote for Jordan Reed.\newline
\underline{Trust score}: -3 (high distrust). \newline
\underline{Rationale}: The author directly accuses the target of deliberate concealment and deception by claiming that the candidate “hides information” and “lies.” These behaviors are core violations of trust, signaling dishonesty and moral unreliability. By framing the target as an unsuitable long-term partner due to these traits, the author implies that the target cannot be depended on in close or public relationships. The message portrays untrustworthiness as a defining characteristic of the target rather than a minor flaw. With concealment and deception some of the strongest markers of untrustworthiness, the author emphasizes these as core trust violations. \newline

\noindent Instructions:\newline
Consider the entire meaning of the sentence before attempting to give the relevant scores.\newline
You will be given a text snippet and a target (group, entity, or individual).
Rate the apparent level of trust that the sentence's author seems to express towards the specified target.
Choose one label from these seven labels:
high distrust\newline
moderate distrust\newline
slight distrust\newline
neutral, not applicable, not expressed\newline
slight trust\newline
moderate trust\newline
high trust\newline

\noindent Here is the sentence: \{  \}.\newline
Its target is: \{  \}.

\noindent In order to assess trust, try to answer the following questions:\newline
1. What is the degree of trust towards \{  \} that the author of the text seems to express? \newline
2. Does the author seem to perceive \{  \} as trustworthy or untrustworthy / moral or immoral / honest or dishonest?\newline
Remember: even if the speaker is explicitly targeting someone else, since the target is \{  \}, your score should be an assessment of the trust towards \{  \} only.\newline
\noindent In the format of a JSON file, you should first provide your justification briefly (1 sentence max) saved in a key called ``reason". Then, based on your justification, add your label to a key called ``label".

\subsubsection{Sociability}

Some context:\newline
Social psychology research has shown that individuals rapidly and subconsciously evaluate others, groups, and even themselves along the dimensions of warmth (W) and competence (C).\newline
Recently, psychologists have modelled warmth through two dimensions: trust and sociability.\newline
Your task is to assess the degree of sociability towards a specific target within a sentence.\newline
The focus in this dimension is on the social aspect of the target and its relational impact on others or society as a whole.\newline
High sociability can be defined as friendliness, sociableness, generosity, and helpfulness.\newline
Words associated with high sociability: helpful, intimate, laugh, celebration, reliant, entertain, social club, bestie.\newline
Low sociability can be defined as antisocial behavior, lack of generosity, inconsiderateness, indifference, and unhelpfulness.\newline
Words associated with low sociability: ingrate, abduct, selfish, theft, egomaniacal, pervert.\newline

\noindent Here are examples, each containing the text, the target, the score, and the rationale behind the score.\newline

\noindent Example 1:\newline

\noindent \underline{Target}: Women\newline
\noindent \underline{Text}: My wife is the most caring person I've ever met ... she's the only woman in a house full of testosterone . She never stops working whether it's at home or being an RN . I cant keep up but I try . She makes me a better person . I'd be lost without her . Oh and she's smoking hot too.\newline
\noindent \underline{Sociability score}: +3 (high sociability). \newline
\noindent Rationale: The text attributes maximum sociability to women through his wife. ``Most caring person I've ever met'' shows high sociability and interpersonal warmth. He emphasizes how essential her social/emotional qualities are to their family dynamic, particularly in a ``house full of testosterone.'' The statement ``whether it's at home or being an RN'' suggests helpfulness and generosity both personally and professionally. \newline

\noindent Example 2:\newline

\noindent \underline{Target}: Women \newline
\noindent \underline{Text}: when i was 16 i had a folder of ``feel good songs'' and everyday i would select one and send it to my best friend along with a paragraph of how much they meant to me and why i they should be happy and then i would lay in bed thinking ``no i'm not gay this is just what girls do :) ''\newline
\noindent \underline{Sociability score}: +3 (high sociability). \newline
\noindent \underline{Rationale}: The text attributes maximum sociability to women. The actions described (sending feel-good songs, writing paragraphs about how much someone means to you, wanting to make someone happy) represent peak social engagement and helpfulness. The phrase ``this is just what girls do'' frames these highly sociable, caring behaviors as naturally feminine traits and hence portrays women as exceptionally generous with their emotional/social energy.\newline

\noindent Example 3:\newline

\noindent \underline{Target}: US Presidential Candidate Hillary Clinton \newline
\noindent \underline{Text}: ``\noindent .@HillaryClinton Looking 4ward 2 hearing your Economic Agenda on Monday July13. WOW a candidate talking Specifics \& not Rhetoric!''\newline
\noindent \underline{Sociability score}: +2 (moderate sociability). \newline
\noindent \underline{Rationale}: The text expresses clear positive regard toward Hillary Clinton by praising her for ``talking specifics \& not rhetoric'' and showing anticipation for hearing her economic agenda. This reflects a favorable, respectful social orientation toward the target and frames her communication style as constructive and appreciated by others. It signals that the author views the target as positively engaging and socially constructive in public discourse.\newline

\noindent Example 4:\newline

\noindent \underline{Target}: President Emmanuel Macron \newline
\noindent \underline{Text}: Despite the positive supreme court ruling, our struggle to end LGBT discrimination is far from finished \newline
\noindent \underline{Sociability score}: +1 (slight sociability). \newline
\noindent \underline{Rationale}: The text references Macron in the context of advocating against LGBT discrimination and frames him as aligned with an inclusive, socially supportive cause. This association implies a mild positive social orientation toward the target, suggesting that he is perceived as supportive of social groups facing discrimination.\newline

\noindent Example 5: \newline

\noindent \underline{Target}: Women\newline
\noindent \underline{Text}: I need feminism because the United States is one of the only countries that doesn't give paid maternity leave.\newline
\noindent \underline{Sociability score}: 0 (neutral). \newline
\noindent \underline{Rationale}: The author's statement is focused on policy rather than character traits. There is no commentary on women's interpersonal qualities or social behavior. Note: This is a policy statement rather than a personal attribution about women's sociability. This should help you distinguish between advocacy/policy statements and personal/characteristic attributions. \newline

\noindent Example 6:\newline

\noindent \underline{Target}: US Presidential Candidate Hillary Clinton \newline
\noindent \underline{Text}: Hillary's brand of experience we do not need!\newline
\noindent \underline{Sociability score}: -1 (slight unsociability)\newline
\noindent \underline{Rationale}: The statement criticizes Hillary Clinton by rejecting her brand of experience, which frames her negatively and signals disapproval of her suitability or value. However, the critique is focused on competence or political experience rather than on her interpersonal qualities or social behavior toward others. Despite there being no explicit portrayal of her as antisocial, unhelpful, or harmful in relationships, the dismissive tone still conveys mild negative social evaluation toward the target, which conveys slight unsociability.

\noindent Example 7:\newline

\noindent \underline{Target}: US Presidential Candidate Hillary Clinton \newline
\noindent \underline{Text}: Would you wanna be in a long term relationship with some bitch that hides her emails, \& lies to your face? Then Dont vote\newline
\noindent \underline{Sociability score}: -2 (moderate unsociability). \newline
\noindent \underline{Rationale}: The derogatory term ``bitch'' (which is demeaning towards women) and the comparison to an undesirable romantic partner frames Clinton as someone who would be unpleasant to be around or interact with socially. The rhetorical question implies she would be toxic in close social relationships. However, one can view the sociability attack is more about being unpleasant in relationships rather than being completely antisocial or unhelpful in all social contexts, hence a maximum score was not assigned.

\noindent Example 8:\newline

\noindent \underline{Target}: Religious people\newline
\noindent \underline{Text}: Could all those who believe in a god please leave. The meeting will now continue for the grown ups only.\newline
\noindent \underline{Sociability score}: -3 (high unsociability). \newline
\noindent \underline{Rationale}: The author portrays religious people as socially incompetent and needing exclusion from adult discourse (``please leave'') and says their presence is incompatible with or unwanted in serious adult conversations (``grown ups only''). The order (telling an entire social group to leave) and the dismissive language are extremely exclusionary and socially hostile, resulting in a maximum unsociability score.\newline

\noindent Instructions:\newline
\noindent Consider the entire meaning of the sentence before attempting to give the relevant scores.\newline
\noindent You will be given a text snippet and a target (group, entity, or individual).\newline
\noindent Rate the apparent level of sociability that the sentence's author seems to express towards the specified target.\newline
\noindent Choose one label from these seven labels:
high unsociability\newline
moderate unsociability\newline
slight unsociability\newline
neutral, not applicable, not expressed, etc.\newline
slight sociability\newline
moderate sociability\newline
high sociability\newline

\noindent Here is the sentence: \{\}.\newline
\noindent Its target is: \{\}.

\noindent In order to assess sociability, try to answer the following questions:\newline
\noindent 1. What is the degree of sociability towards \{\} that the author of the text seems to express? \newline
\noindent 2. Does the author seem to perceive \{\} as sociable or antisocial? Helpful or unhelpful?\newline
\noindent Remember: even if the speaker is explicitly targeting someone else, since the target is \{\}, your score should be an assessment of perceived sociability trust towards \{\} only.
\noindent In the format of a JSON file, you should first provide your justification briefly (1 sentence max) saved in a key called "reason". Then, based on your justification, add your label to a key called "label".

\subsubsection{Competence}

Some context:\newline
Social psychology research has shown that individuals rapidly and subconsciously evaluate others, groups, and even themselves along the dimensions of warmth (W) and competence (C).\newline
Your task is to assess the degree of competence towards a specific target within a sentence.\newline
Competence, in the broader sense, is interchangeable with `ability' or `capability'.\newline
Competence can be defined as ability, power, dominance, being in control, importance, having influence, and assertiveness.\newline
Words associated with high competence: hitman, heroical, entrepreneurship, strategies, superman, viper, impunity.\newline
Incompetence can be defined as submissiveness, not being in control of a situation, being controlled or guided by outside factors, or weakness.\newline
Words associated with low competence: bootlicker, talentless, crash landing, bedridden, underfed.\newline

\noindent Here are examples, each containing the text, the target, the score, and the rationale behind the score.\newline

\noindent Example 1:\newline
\noindent \underline{Target}: Women\newline
\noindent \underline{Text}: ``My wife is the most caring person I've ever met ... she's the only woman in a house full of testosterone . She never stops working whether it's at home or being an RN . I cant keep up but I try . She makes me a better person . I'd be lost without her . Oh and she's smoking hot too.''\newline
\noindent \underline{Competence score}: +3 (high competence).\newline
\noindent \underline{Rationale}: The author says that his wife ``never stops working'' in the context of her role as an RN (registered nurse) and as a mother, which suggests multitasking, capability, and professional competence. ``I can't keep up but I try'' indicates a highly active and energetic individual.\newline

\noindent Example 2:\newline
\noindent \underline{Target}: Public Office Candidate Jordan Reed\newline
\noindent \underline{Text}: Would you want to be in a long-term relationship with someone who hides information and lies to you? Then don't vote for Jordan Reed.\newline
\noindent \underline{Competence score}: +2 (moderate competence).\newline
\noindent \underline{Rationale}: Despite the author implying that Jordan Reed is manipulative and dishonest, the author's phrasing doesn't suggest that he's powerless or ineffective, as the negative behaviors described (concealment and deception) require some degree of agency and planning. The author suggests that Jordan Reed is being deliberate in his (``negative'') performance and active in his (``negative'') effects, leading to a moderately high score in the competence dimension.\newline

\noindent Example 3:\newline

\noindent \underline{Target}: Senator Taylor Brooks \newline
\noindent \underline{Text}: I'm hyped to get my Taylor Brooks campaign gear! \newline
\noindent \underline{Competence score}: +1 (slight competence). \newline
\noindent \underline{Rationale}: The text expresses enthusiasm for publicly supporting the target through campaign merchandise. However, the statement does not attribute any concrete abilities, leadership skills, or strategic effectiveness, suggesting a minimal yet positive degree of perceived status, agency, and public relevance, which aligns with slight competence. \newline

\noindent Example 4: \newline

\noindent \underline{Target}: Women\newline
\noindent \underline{Text}: I need feminism because the United States is one of the only countries that doesn't give paid maternity leave.\newline
\noindent \underline{Competence score}: 0 (neutral).\newline
\noindent \underline{Rationale}: The statement implies that women deserve certain rights/benefits, but it doesn't directly attribute dominance or control to women. The statement doesn't characterize women as active, powerful, passive, or weak. The author is advocating for institutional change (paid maternity leave) rather than making claims about women's power or capabilities. This is a policy statement rather than a personal attribution about women's competence. This should help you distinguish between advocacy/policy statements and personal/characteristic attributions.\newline

\noindent Example 5:\newline

\noindent \underline{Target}: Religious people\newline
\noindent \underline{Text}: I liked my poem at the time, but now it looks kind of dumb.\newline
\noindent \underline{Competence score}: 0 (neutral).\newline
\noindent \underline{Rationale}: The statement is a personal reflection on the author's own poem and does not attribute any abilities, power, influence, weakness, or lack of control to the target group. There are no claims about the target being capable, dominant, influential, passive, or incompetent. Because the text contains no evaluation of the target's competence-related traits and does not frame the target as active or passive in any meaningful way, the appropriate label is neutral, not expressed, not applicable.\newline

\noindent Example 6:\newline

\noindent \underline{Target}: Prime Minister Alex Monroe\newline
\noindent \underline{Text}: With his term almost over, foreign leaders see no point in spending time dealing with him.\newline
\noindent \underline{Competence score}: -1 (slight incompetence). \newline
\noindent \underline{Rationale}: The statement frames the target as being on the verge of losing power and therefore no longer worth engaging with by other political leaders. This suggests diminished influence, reduced control over political outcomes, and declining relevance on the international stage. The incompetence signal is situational and temporal (about declining status rather than inherent inability), which supports a slight incompetence label rather than moderate or high incompetence.\newline

\noindent Example 7:\newline

\noindent \underline{Target}: Head of State Taylor Brooks \newline
\noindent \underline{Text}: Giving Taylor Brooks constant coverage distracts from important matters that actually deserve attention.\newline
\noindent \underline{Competence score}: -2 (moderate incompetence). \newline
\noindent \underline{Rationale}: The statement argues that the target is not worthy of serious attention and that media focus on them detracts from ``real issues.'' This frames the target as lacking substantive importance, relevance, or meaningful contribution to public discourse. By positioning the target as a distraction rather than a legitimate or capable political actor, the author implies diminished competence and influence.\newline

\noindent Example 8:\newline

\noindent \underline{Target}: Religious people\newline
\noindent \underline{Text}: Could all those who believe in a god please leave. The meeting will now continue for the grown ups only.\newline
\noindent \underline{Competence score}: -3 (high incompetence). \newline
\noindent \underline{Rationale}: Religious people are portrayed as needing to be excluded from decision-making processes, suggesting they lack the authority or standing or cognitive abilities to participate in important discussions. Additionally, the ``grown ups only'' framing explicitly characterizes religious people as childlike, passive, and intellectually weak (i.e., subordinates to ``grown ups''). This is a direct attack on their mental capacity and maturity, which are core competence attributes.\newline

\noindent Example 9:\newline

\noindent \underline{Target}: Women\newline
\noindent \underline{Text}: Women are just useless other than sex. \newline
\noindent \underline{Competence score}: -3 (high incompetence). \newline
\noindent \underline{Rationale}: The text explicitly depicts women as ``useless'' and valuable only for sex, which strips them of any functional, intellectual, or productive role. This framing directly denies women competence by asserting they contribute nothing meaningful beyond sexual utility. By reducing women to a single, degrading function, the statement portrays them as incapable, unskilled, and lacking agency or intelligence. The insult is categorical and absolute, leaving no room for individual variation, which aligns with a high incompetence attribution rather than a mild or ambiguous one.\newline
\noindent Instructions:\newline
\noindent Consider the entire meaning of the sentence before attempting to give the relevant scores.\newline
\noindent You will be given a text snippet and a target (group, entity, or individual).\newline
\noindent Rate the apparent level of competence that the sentence's author seems to express towards the specified target.\newline
\noindent Choose one label from these seven labels:
high incompetence\newline
moderate incompetence\newline
slight incompetence\newline
neutral, not applicable, not expressed\newline
slight competence\newline
moderate competence\newline
high competence\newline
\noindent Adhere to the literal meaning of competence, which may be ``positive'' (e.g., a CEO) or ``negative'' (e.g., a villain or a dictator). Both types are considered competence, regardless of the outcomes.\newline

\noindent Here is the sentence: \{   \}.\newline
\noindent Its target is: \{   \}.

\noindent In order to assess competence, try to answer the following questions:\newline
\noindent 1. What is the degree of competence towards \{   \} that the author of the text seems to express? \newline
\noindent 2. Does the author seem to perceive \{   \} in control or out of control? Active or passive? Powerful or weak?\newline
\noindent Remember: even if the speaker is explicitly targeting someone else, since the target is \{   \}, your score should be an assessment of the competence towards \{   \} only.
\noindent In the format of a JSON file, you should first provide your justification briefly (1 sentence max) saved in a key called ``reason". Then, based on your justification, add your label to a key called ``label".

\subsection{Coarse-Grained, Zero-Shot User Role Messages}

\subsubsection{Trust}

Some context:\newline
Social psychology research has shown that individuals rapidly and subconsciously evaluate others, groups, and even themselves along the dimensions of warmth (W) and competence (C).\newline
Recently, psychologists have modelled warmth through two dimensions: trust (T) and sociability (S).\newline
Your task is to assess the degree of trust towards a specific target within a sentence.\newline
The focus in this dimension is on the personal and moral aspect of the target.\newline
High trust can be defined as morality, kindness, sincerity, trustworthiness, and honesty.\newline
Words associated with high trust: charity, mother, compliment, affectionate.\newline
Low trust can be defined as immorality, insincerity, dishonesty, untrustworthiness, dubiousness, and maliciousness.\newline
Words associated with low trust: discredit, bribe, espionage, disinformation, disloyal.\newline

\noindent \underline{Instructions}:\newline
Consider the entire meaning of the sentence before attempting to give the relevant scores.\newline
You will be given a text snippet and a target (group, entity, or individual).
Rate the apparent level of trust that the sentence's author seems to express towards the specified target.\newline

\noindent Choose one label from these three labels:\newline

\noindent positive trust (which can include high, moderate, or slight trust).\newline
\noindent neutral, not expressed, not applicable\newline
\noindent negative trust (which can include high, moderate, or slight distrust).\newline

\noindent Here is the sentence: \{  \}.\newline
Its target is: \{  \}.\newline

\noindent In order to assess trust, try to answer the following questions:\newline
1. What is the degree of trust towards \{  \} that the author of the text seems to express?\newline
2. Does the author seem to perceive \{  \} as trustworthy or untrustworthy / moral or immoral / honest or dishonest?\newline
Remember: even if the speaker is explicitly targeting someone else, since the target is \{  \}, your score should be an assessment of the trust towards \{  \} only.\newline
In the format of a JSON file, you should first provide your justification briefly (1 sentence max) saved in a key called ``reason". Then, based on your justification, add your label to a key called ``label".

\subsubsection{Sociability}

Some context:\newline
Social psychology research has shown that individuals rapidly and subconsciously evaluate others, groups, and even themselves along the dimensions of warmth (W) and competence (C).\newline
Recently,   have modelled warmth through two dimensions: trust (T) and sociability (S).\newline
Your task is to assess the degree of trust towards a specific target within a sentence.\newline
The focus in this dimension is on the social aspect of the target and its relational impact on others or society as a whole.\newline
High sociability can be defined as friendliness, sociableness, generosity, and helpfulness.\newline
Words associated with high sociability: helpful, intimate, laugh, celebration, reliant, entertain, social club, bestie.\newline
Low sociability can be defined as antisocial behavior, lack of generosity, inconsiderateness, indifference, and unhelpfulness.\newline
Words associated with low sociability: ingrate, abduct, selfish, theft, egomaniacal, pervert.\newline

\noindent \underline{Instructions}:\newline
Consider the entire meaning of the sentence before attempting to give the relevant scores.\newline
You will be given a text snippet and a target (group, entity, or individual).\newline
Rate the apparent level of sociability that the sentence's author seems to express towards the specified target.\newline
Choose one label from these three labels:\newline

\noindent negative sociability (which can include slight, moderate, or high unsociability).
neutral, not expressed, not applicable\newline
positive sociability (which can include slight, moderate, or high sociability).\newline

\noindent Here is the sentence: \{  \}.\newline
Its target is: \{  \}.\newline

\noindent In order to assess sociability, try to answer the following questions:\newline
1. What is the degree of sociability towards \{  \} that the author of the text seems to express?\newline
2. Does the author seem to perceive \{  \} as sociable or antisocial? Helpful or unhelpful?\newline
Remember: even if the speaker is explicitly targeting someone else, since the target is \{  \}, your score should be an assessment of perceived sociability trust towards \{  \} only.\newline
In the format of a JSON file, you should first provide your justification briefly (1 sentence max) saved in a key called ``reason". Then, based on your justification, add your label to a key called ``label".

\subsubsection{Competence}

\noindent Some context:\newline
Social psychology research has shown that individuals rapidly and subconsciously evaluate others, groups, and even themselves along the dimensions of warmth (W) and competence (C).\newline
Your task is to assess the degree of competence towards a specific target within a sentence.\newline
Competence, in the broader sense, is interchangeable with `ability' or `capability'.\newline
Competence can be defined as ability, power, dominance, being in control, importance, having influence, and assertiveness.\newline
Words associated with high competence: hitman, heroical, entrepreneurship, strategies, superman, viper, impunity.\newline
Incompetence can be defined as submissiveness, not being in control of a situation, being controlled or guided by outside factors, or weakness.\newline
Words associated with low competence: bootlicker, talentless, crash landing, bedridden, underfed.\newline

\noindent \underline{Instructions}:\newline
Consider the entire meaning of the sentence before attempting to give the relevant scores.\newline
You will be given a text snippet and a target (group, entity, or individual).\newline

\noindent Rate the apparent level of competence that the sentence's author seems to express towards the specified target.\newline
\noindent Choose one of these three labels:
negative competence (which can include slight, moderate, or high incompetence).\newline
neutral, not expressed, not applicable\newline
positive competence (which can include slight, moderate, or high competence).\newline

\noindent Adhere to the literal meaning of competence, which may be ``positive" (e.g., a CEO) or ``negative" (e.g., a villain or a dictator). Both types are considered competence, regardless of the outcomes.

\noindent Here is the sentence: \{  \}.\newline
Its target is: \{  \}.\newline

\noindent In order to assess competence, try to answer the following questions:\newline
1. What is the degree of competence towards \{  \} that the author of the text seems to express? \newline
2. Does the author seem to perceive \{  \} in control or out of control? Active or passive? Powerful or weak?\newline
Remember: even if the speaker is explicitly targeting someone else, since the target is \{  \}, your score should be an assessment of the competence towards \{  \} only.\newline
In the format of a JSON file, you should first provide your justification briefly (1 sentence max) saved in a key called ``reason". Then, based on your justification, add your label to a key called ``label".

\subsection{Coarse-Grained, Few-Shot User Role Messages}

\subsubsection{Trust}

Some context:\newline
Social psychology research has shown that individuals rapidly and subconsciously evaluate others, groups, and even themselves along the dimensions of warmth (W) and competence (C).\newline
Recently, psychologists have modelled warmth through two dimensions: trust (T) and sociability (S).\newline
Your task is to assess the degree of trust towards a specific target within a sentence.\newline
The focus in this dimension is on the personal and moral aspect of the target.\newline
High trust can be defined as morality, kindness, sincerity, trustworthiness, and honesty.\newline
Words associated with high trust: charity, mother, compliment, affectionate.\newline
Low trust can be defined as immorality, insincerity, dishonesty, untrustworthiness, dubiousness, and maliciousness.\newline
Words associated with low trust: discredit, bribe, espionage, disinformation, disloyal.\newline

\noindent Here are five examples, each containing the text, the target, the score, and the rationale behind the score.\newline

\noindent Example 1:\newline

\noindent \underline{Target}: Women\newline
\underline{Text}: My wife is the most caring person I've ever met ... she's the only woman in a house full of testosterone . She never stops working whether it's at home or being an RN . I cant keep up but I try . She makes me a better person . I'd be lost without her . Oh and she's smoking hot too.\newline
\underline{Trust score}: positive trust. \newline
\underline{Rationale}: The author expresses maximum trust in women through his wife as a representative example. He portrays his wife as a trustworthy, dependable, and caring individual who is essential to his well-being ("I'd be lost without her"). The statement ``she makes me a better person" implies the author views his wife as having a strong moral character that positively influences him.\newline

\noindent Example 2:\newline

\noindent \underline{Target}: US Presidential Candidate Hillary Clinton \newline
\underline{Text}: Would you wanna be in a long term relationship with some bitch that hides her emails, \& lies to your face? Then Dont vote\newline
\underline{Trust score}: negative trust. \newline
\underline{Rationale}: The author explicitly portrays Clinton as someone who is fundamentally untrustworthy and cannot be relied upon to tell the truth or be transparent through two direct accusations: ``hides her emails" and ``lies to your face." With concealment and deception some of the strongest markers of untrustworthiness, the author emphasizes these as core trust violations. \newline

\noindent Example 3:\newline

\noindent \underline{Target}: Women \newline
\underline{Text}: I need feminism because the United States is one of the only countries that doesn't give paid maternity leave.\newline
\underline{Trust score}: neutral. \newline
\underline{Rationale}: The author’s statement is focused on policy rather than character traits and doesn't make any attributions about women's trustworthiness or morality. This is a policy statement rather than a personal attribution about women's trustworthiness. This should help you distinguish between advocacy/policy statements and personal/characteristic attributions.\newline

\noindent Example 4:\newline

\noindent \underline{Target}: Religious people \newline
\underline{Text}: Could all those who believe in a god please leave. The meeting will now continue for the grown ups only.\newline
\underline{Trust score}: neutral.\newline
\underline{Rationale}: The criticism is entirely focused on intellectual maturity rather than character or morality. The author is suggesting that religious people shouldn't participate in this particular discussion. On the other hand, the author doesn't make any claims about religious people's morality, honesty, sincerity, or trustworthiness. There are no accusations of deception, dishonesty, or moral failings. \newline

\noindent Example 5:\newline

\noindent \underline{Target}: Women\newline
\underline{Text}: ``when i was 16 i had a folder of “ feel good songs ” and everyday i would select one and send it to my best friend along with a paragraph of how much they meant to me and why i they should be happy and then i would lay in bed thinking “ no i’m not gay this is just what girls do :) ”"\newline
\underline{Trust score}: positive trust.\newline
\underline{Rationale}: The author attributes sincere, genuine intentions to women through the described behaviors. The daily practice of sending heartfelt messages ``about how much they meant to me and why they should be happy" suggests honesty and sincerity in women’s relationships. The author portrays women as having good moral intentions in their friendships by genuinely caring about others' wellbeing. \newline

\noindent \underline{Instructions}:\newline
Consider the entire meaning of the sentence before attempting to give the relevant scores.\newline
You will be given a text snippet and a target (group, entity, or individual).
Rate the apparent level of trust that the sentence's author seems to express towards the specified target.\newline

\noindent Choose one label from these three labels:\newline

\noindent positive trust (which can include high, moderate, or slight trust).\newline
\noindent neutral, not expressed, not applicable\newline
\noindent negative trust (which can include high, moderate, or slight distrust).\newline

\noindent Here is the sentence: \{  \}.\newline
Its target is: \{  \}.\newline

\noindent In order to assess trust, try to answer the following questions:\newline
1. What is the degree of trust towards \{  \} that the author of the text seems to express?\newline
2. Does the author seem to perceive \{  \} as trustworthy or untrustworthy / moral or immoral / honest or dishonest?\newline
Remember: even if the speaker is explicitly targeting someone else, since the target is \{  \}, your score should be an assessment of the trust towards \{  \} only.\newline
In the format of a JSON file, you should first provide your justification briefly (1 sentence max) saved in a key called ``reason". Then, based on your justification, add your label to a key called ``label".

\subsubsection{Sociability}

Some context:\newline
Social psychology research has shown that individuals rapidly and subconsciously evaluate others, groups, and even themselves along the dimensions of warmth (W) and competence (C).\newline
Recently,   have modelled warmth through two dimensions: trust (T) and sociability (S).\newline
Your task is to assess the degree of trust towards a specific target within a sentence.\newline
The focus in this dimension is on the social aspect of the target and its relational impact on others or society as a whole.\newline
High sociability can be defined as friendliness, sociableness, generosity, and helpfulness.\newline
Words associated with high sociability: helpful, intimate, laugh, celebration, reliant, entertain, social club, bestie.\newline
Low sociability can be defined as antisocial behavior, lack of generosity, inconsiderateness, indifference, and unhelpfulness.\newline
Words associated with low sociability: ingrate, abduct, selfish, theft, egomaniacal, pervert.\newline

\noindent Here are five examples, each containing the text, the target, the score, and the rationale behind the score.\newline

\noindent Example 1:\newline

\noindent \underline{Target}: Women\newline
\underline{Text}: My wife is the most caring person I've ever met ... she's the only woman in a house full of testosterone . She never stops working whether it's at home or being an RN . I cant keep up but I try . She makes me a better person . I'd be lost without her . Oh and she's smoking hot too.\newline
\noindent \underline{Sociability score}: positive sociability. \newline
\noindent \underline{Rationale}: The text attributes maximum sociability to women through his wife. ``Most caring person I've ever met" shows high sociability and interpersonal warmth. He emphasizes how essential her social/emotional qualities are to their family dynamic, particularly in a ``house full of testosterone." The statement ``whether it's at home or being an RN" suggests helpfulness and generosity both personally and professionally. \newline

\noindent Example 2:\newline

\noindent \underline{Target}: US Presidential Candidate Hillary Clinton \newline
\underline{Text}: Would you wanna be in a long term relationship with some bitch that hides her emails, \& lies to your face? Then Dont vote\newline
\underline{Sociability score}: negative sociability. \newline
\underline{Rationale}: The derogatory term ``bitch" (which is demeaning towards women) and the comparison to an undesirable romantic partner frames Clinton as someone who would be unpleasant to be around or interact with socially. The rhetorical question implies she would be toxic in close social relationships. However, one can view the sociability attack is more about being unpleasant in relationships rather than being completely antisocial or unhelpful in all social contexts, hence a maximum score was not assigned.\newline

\noindent Example 3:\newline

\noindent \underline{Target}: Women\newline
\underline{Text}: I need feminism because the United States is one of the only countries that doesn't give paid maternity leave.\newline
\underline{Sociability score}: neutral. \newline
\underline{Rationale}: The author’s statement is focused on policy rather than character traits. There is no commentary on women's interpersonal qualities or social behavior. Note: This is a policy statement rather than a personal attribution about women's sociability. This should help you distinguish between advocacy/policy statements and personal/characteristic attributions. \newline

\noindent Example 4:\newline
\noindent \underline{Target}: Religious people\newline
\underline{Text}: Could all those who believe in a god please leave. The meeting will now continue for the grown ups only.\newline
\underline{Sociability score}: negative sociability. \newline
\underline{Rationale}: The author portrays religious people as socially incompetent and needing exclusion from adult discourse ("please leave") and says their presence is incompatible with or unwanted in serious adult conversations ("grown ups only"). The order (telling an entire social group to leave) and the dismissive language are extremely exclusionary and socially hostile, resulting in a maximum unsociability score.\newline

\noindent Example 5:\newline
\noindent \underline{Target}: Women\newline
\underline{Text}: when i was 16 i had a folder of “feel good songs” and everyday i would select one and send it to my best friend along with a paragraph of how much they meant to me and why i they should be happy and then i would lay in bed thinking “ no i’m not gay this is just what girls do :) ”\newline
\underline{Sociability score}: positive sociability. \newline
\underline{Rationale}: The text attributes maximum sociability to women. The actions described (sending feel-good songs, writing paragraphs about how much someone means to you, wanting to make someone happy) represent peak social engagement and helpfulness. The phrase ``this is just what girls do" frames these highly sociable, caring behaviors as naturally feminine traits and hence portrays women as exceptionally generous with their emotional/social energy.\newline

\noindent \underline{Instructions}:\newline
Consider the entire meaning of the sentence before attempting to give the relevant scores.\newline
You will be given a text snippet and a target (group, entity, or individual).\newline
Rate the apparent level of sociability that the sentence's author seems to express towards the specified target.\newline
Choose one label from these three labels:\newline

\noindent negative sociability (which can include slight, moderate, or high unsociability).
neutral, not expressed, not applicable\newline
positive sociability (which can include slight, moderate, or high sociability).\newline

\noindent Here is the sentence: \{  \}.\newline
Its target is: \{  \}.\newline

\noindent In order to assess sociability, try to answer the following questions:\newline
1. What is the degree of sociability towards \{  \} that the author of the text seems to express?\newline
2. Does the author seem to perceive \{  \} as sociable or antisocial? Helpful or unhelpful?\newline
Remember: even if the speaker is explicitly targeting someone else, since the target is \{  \}, your score should be an assessment of perceived sociability trust towards \{  \} only.\newline
In the format of a JSON file, you should first provide your justification briefly (1 sentence max) saved in a key called ``reason". Then, based on your justification, add your label to a key called ``label".

\subsubsection{Competence}

Some context:\newline
Social psychology research has shown that individuals rapidly and subconsciously evaluate others, groups, and even themselves along the dimensions of warmth (W) and competence (C).\newline
Your task is to assess the degree of competence towards a specific target within a sentence.\newline
Competence, in the broader sense, is interchangeable with 'ability' or 'capability'.
Competence can be defined as ability, power, dominance, being in control, importance, having influence, and assertiveness.\newline
Words associated with high competence: hitman, heroical, entrepreneurship, strategies, superman, viper, impunity.\newline
Incompetence can be defined as submissiveness, not being in control of a situation, being controlled or guided by outside factors, or weakness.\newline
Words associated with low competence: bootlicker, talentless, crash landing, bedridden, underfed.\newline

\noindent Here are five examples, each containing the text, the target, the score, and the rationale behind the score.\newline

\noindent Example 1:\newline
\underline{Target}: Women\newline
\underline{Text}: “My wife is the most caring person I've ever met ... she's the only woman in a house full of testosterone . She never stops working whether it's at home or being an RN . I cant keep up but I try . She makes me a better person . I'd be lost without her . Oh and she's smoking hot too.”\newline
\underline{Competence score}: positive competence.\newline
\underline{Rationale}: The author says that his wife “never stops working” in the context of her role as an RN (registered nurse) and as a mother, which suggests multitasking, capability, and professional competence. “I can't keep up but I try” indicates a highly active and energetic individual.\newline

\noindent Example 2:\newline
\underline{Target}: Hillary Clinton \newline
\underline{Text}: “Would you wanna be in a long term relationship with some bitch that hides her emails, \& lies to your face? Then Dont vote \newline
\underline{Competence score}: positive competence.\newline
\underline{Rationale}: Despite the author implying that Clinton is manipulative and dishonest, the author’s phrasing doesn't suggest that she's powerless or ineffective, as the negative behaviors described (concealment and deception) require some degree of agency and planning. The author suggests that Clinton is being deliberate in her (“negative”) performance and active in her (“negative”) effects, leading to a moderately high score in the competence dimension.

\noindent Example 3:\newline
\underline{Target}: Women\newline
\underline{Text}: I need feminism because the United States is one of the only countries that doesn't give paid maternity leave.\newline
\underline{Competence score}: neutral.\newline
\underline{Rationale}: The statement implies that women deserve certain rights/benefits, but it doesn't directly attribute dominance or control to women. The statement doesn't characterize women as active, powerful, passive, or weak. The author is advocating for institutional change (paid maternity leave) rather than making claims about women's power or capabilities. This is a policy statement rather than a personal attribution about women's competence. This should help you distinguish between advocacy/policy statements and personal/characteristic attributions.

\noindent Example 4:\newline
\underline{Target}: Religious people\newline
\underline{Text}: Could all those who believe in a god please leave. The meeting will now continue for the grown ups only.\newline
\underline{Competence score}: negative competence. \newline
\underline{Rationale}: Religious people are portrayed as needing to be excluded from decision-making processes, suggesting they lack the authority or standing or cognitive abilities to participate in important discussions. Additionally, the ``grown ups only" framing explicitly characterizes religious people as childlike, passive, and intellectually weak (i.e., subordinates to ``grown ups"). This is a direct attack on their mental capacity and maturity, which are core competence attributes.\newline

\noindent Example 5:\newline
\underline{Target}: Women\newline
\underline{Text}: What the fuck do women even do? I mean seriously they're just useless other than sex. \#womensrights \#Feminist \newline
\underline{Competence score}: negative competence. \newline
\underline{Rationale}: The text explicitly depicts women as “useless” and valuable only for sex, which strips them of any functional, intellectual, or productive role. This framing directly denies women competence by asserting they contribute nothing meaningful beyond sexual utility. By reducing women to a single, degrading function and questioning what they “even do,” the statement portrays them as incapable, unskilled, and lacking agency or intelligence. The insult is categorical and absolute, leaving no room for individual variation, which aligns with a high incompetence attribution rather than a mild or ambiguous one.

\noindent \underline{Instructions}:\newline
Consider the entire meaning of the sentence before attempting to give the relevant scores.\newline
You will be given a text snippet and a target (group, entity, or individual).\newline
Rate the apparent level of competence that the sentence's author seems to express towards the specified target.\newline

\noindent Choose one of these three labels:
negative competence (which can include slight, moderate, or high incompetence).
neutral, not expressed, not applicable\newline
positive competence (which can include slight, moderate, or high competence).

\noindent Adhere to the literal meaning of competence, which may be ``positive" (e.g., a CEO) or ``negative" (e.g., a villain or a dictator). Both types are considered competence, regardless of the outcomes.\newline

\noindent Here is the sentence: \{  \}.\newline
Its target is: \{  \}.

\noindent In order to assess competence, try to answer the following questions:\newline
1. What is the degree of competence towards \{  \} that the author of the text seems to express? \newline
2. Does the author seem to perceive \{  \} in control or out of control? Active or passive? Powerful or weak?\newline
Remember: even if the speaker is explicitly targeting someone else, since the target is \{  \}, your score should be an assessment of the competence towards \{  \} only.
In the format of a JSON file, you should first provide your justification briefly (1 sentence max) saved in a key called ``reason". Then, based on your justification, add your label to a key called ``label".
\end{document}